\documentclass[twoside]{article}

\usepackage[accepted]{aistats2021}
\usepackage[round]{natbib}
\renewcommand{\bibname}{References}
\renewcommand\refname{References}
\renewcommand{\bibsection}{\subsubsection*{\bibname}}

\usepackage[utf8]{inputenc} 
\usepackage[T1]{fontenc}    

\usepackage{booktabs}       
\usepackage{amsfonts}       
\usepackage{nicefrac}       
\usepackage{microtype}      

\usepackage{graphicx}
\usepackage{subfigure}

\usepackage{url}

\usepackage{todonotes}
\usepackage{algorithm,algorithmic}
\usepackage{amsmath}
\usepackage{amssymb}
\usepackage{caption}
\usepackage{amsthm}
\usepackage{bbm}
\usepackage{array,multirow}
\usepackage{wrapfig}
\usepackage[normalem]{ulem}

\usepackage[english]{babel}
\addto{\captionsenglish}{%
    \renewcommand{\bibname}{References}
}
\usepackage{mdframed}   

\definecolor{mydarkblue}{rgb}{0,0.08,0.45}
\usepackage[colorlinks,citecolor=mydarkblue,urlcolor=mydarkblue,linkcolor=mydarkblue,filecolor=mydarkblue]{hyperref}

\usepackage{amsmath,amsfonts,bm}
\usepackage{upgreek}

\def\eqref#1{equation~\ref{#1}}

\def\1{\bm{1}}

\def\rtau{{\uptau}}
\def\rtheta{{\uptheta}}
\def\rlambda{{\uplambda}}
\def\rbeta{{\upbeta}}
\def\rkappa{{\upkappa}}

\def\rf{{\textnormal{f}}}

\def\rh{{\textnormal{h}}}
\def\ri{{\textnormal{i}}}

\def\rw{{\textnormal{w}}}
\def\rx{{\textnormal{x}}}
\def\ry{{\textnormal{y}}}
\def\rz{{\textnormal{z}}}

\def\rvtheta{{\boldsymbol{\uptheta}}}

\def\rvlambda{{\boldsymbol{\uplambda}}}

\def\rvbeta{{\boldsymbol{\upbeta}}}

\def\rvf{{\mathbf{f}}}

\def\rvh{{\mathbf{h}}}

\def\rvy{{\mathbf{y}}}

\def\rmF{{\mathbf{F}}}

\def\rmW{{\mathbf{W}}}
\def\rmX{{\mathbf{X}}}
\def\rmY{{\mathbf{Y}}}

\def\rmXi{{\boldsymbol{\Upxi}}}

\def\vmu{{\bm{\mu}}}
\def\vtheta{{\bm{\theta}}}
\def\vlambda{{\bm{\lambda}}}

\def\vphi{{\bm{\phi}}}

\def\vh{{\bm{h}}}

\def\vw{{\bm{w}}}
\def\vx{{\bm{x}}}
\def\vy{{\bm{y}}}

\def\mF{{\bm{F}}}

\def\mW{{\bm{W}}}
\def\mX{{\bm{X}}}
\def\mY{{\bm{Y}}}

\def\mPhi{{\bm{\Phi}}}

\def\mSigma{{\bm{\Sigma}}}

\def\mXi{{\bm{\Xi}}}

\DeclareMathAlphabet{\mathsfit}{\encodingdefault}{\sfdefault}{m}{sl}
\SetMathAlphabet{\mathsfit}{bold}{\encodingdefault}{\sfdefault}{bx}{n}

\newcommand{\minus}{\scalebox{0.75}[1.0]{$-$}}
\usepackage[capitalize]{cleveref}

\begin{document}

%

%

\twocolumn[

\aistatstitle{Predictive Complexity Priors}

\aistatsauthor{Eric Nalisnick \And Jonathan Gordon \And José Miguel Hernández-Lobato}

\aistatsaddress{ University of Amsterdam \And  University of Cambridge \And University of Cambridge } ]

\begin{abstract}
  Specifying a Bayesian prior is notoriously difficult for complex models such as neural networks.  Reasoning about parameters is made challenging by the high-dimensionality and over-parameterization of the space.  Priors that seem benign and uninformative can have unintuitive and detrimental effects on a model's predictions.  For this reason, we propose \textit{predictive complexity priors}: a functional prior that is defined by comparing the model's predictions to those of a reference model.  Although originally defined on the model outputs, we transfer the prior to the model parameters via a change of variables.  The traditional Bayesian workflow can then proceed as usual.  We apply our predictive complexity prior to high-dimensional regression, reasoning over neural network depth, and sharing of statistical strength for few-shot learning.
\end{abstract}

\section{INTRODUCTION}
Choosing the prior for a Bayesian model is the most important---and often, the most difficult---step in model specification \citep{robert2007bayesian}.  Unfortunately, prior specification within machine learning is additionally fraught and challenging.  Popular models such as neural networks (NNs) are high dimensional and unidentifiable, making it extremely hard to reason about what makes a good prior.  Moreover, since the true posterior can almost never be recovered, it is difficult to isolate a prior's influence (even empirically).  We are left asking: do the specifics of the prior even matter if they are blunted by our posterior approximations and large data sets?  Until recently, most work in machine learning has assumed the negative and resorted to priors of convenience.  For instance, the standard normal distribution is by far the most popular prior for Bayesian NNs \citep{Zhang2020Cyclical, heek2019bayesian, wenzel2020good}. 


In this paper, we present a novel framework to specify priors for black-box models.  Rather than working with the uninterpretable parameter space, we place the Bayesian prior on the model's \emph{functional complexity}.  Our prior, termed the \textit{predictive complexity prior} (PredCP), compares the model's predictions to those of a reference model.  For example, we define the reference model for a NN to be a NN with one fewer layer.  The PredCP can then assess and control the effect of depth on the model's capacity.  Unlike previous work on functional priors \citep{sun2018functional}, we use a change of variables to \emph{exactly} translate the functional prior into a proper prior on the model parameters.  Bayesian inference can then proceed as usual and without involving extra machinery.  We claim the following contributions:
\begin{itemize}
    \item \textbf{Methodology:} We propose \textit{predictive complexity priors} (PredCPs).  These extend \citet{simpson2017penalising}'s framework to the model predictions, thereby allowing our data-space intuitions to guide prior specification.  Moreover, we introduce crucial modifications that allow the PredCP to scale to large, black-box models such as NNs. 
    \item \textbf{Applications:}  We demonstrate the wide applicability of the PredCP by using it for three disparate tasks: high-dimensional regression, reasoning over depth in Bayesian NNs, and sharing information across tasks for few-shot learning.  For Bayesian NNs, we investigate the PredCP's behavior in detail, revealing its mechanism of action: regularizing predictive variance.
    \item \textbf{Experiments:}  We report results across a variety of tasks (classification, regression, few-shot learning), models (logistic regression, NNs), and posterior inference strategies (Markov chain Monte Carlo, variational inference, MAP estimation).  The PredCP provides consistent improvements in predictive generalization over alternative priors (uninformative, shrinkage).
\end{itemize}

\section{SETTING OF INTEREST}\label{sec:soi}
\paragraph{Notation} Matrices are denoted with upper-case and bold letters (e.g.\ $\mY$), vectors with lower-case and bold (e.g.\ $\vy$), and scalars with no bolding (e.g.\ $y$ or $Y$).  We use italics to differentiate observations and constants (e.g.~$\vy$, $\vtheta$) from random variables (e.g.~$\rvy$, $\rvtheta$).

\paragraph{Model} We consider hierarchical models of the form: \begin{equation}\label{eq:model}
    \rvy \sim p(\rvy | \rvtheta), \ \ \  \ \rvtheta \sim p(\rvtheta | \rtau), \ \ \ \ \rtau \sim p(\rtau)
\end{equation} where $p(\rvy | \rvtheta)$ is the data (sampling) model, $p(\rvtheta | \rtau)$ is a first-level prior, and $p(\rtau)$ is a second-level hyper-prior.  We are primarily concerned with models for which $p(\rvy | \rvtheta)$ is parameterized by a complicated function and $\rtau$ plays a significant role in controlling the complexity of that function.  One such example is parameterizing $\rvy | \rvtheta$ with a NN whose weights are given a normal prior with variance $\rtau$: $\rvtheta \sim \text{N}(\mathbf{0}, \rtau \mathbb{I})$.  As $\rtau$ grows, the weights become less constrained and the model becomes more flexible.  A common strategy for controlling $\rtau$ is to give it a shrinkage prior such as a zero-favoring inverse gamma \citep{neal1994bayesian} or half-Cauchy \citep{carvalho2009handling}.  While this is a sensible approach, it can be hard to understand how $p(\rtau)$ regularizes the downstream predictive function \citep{piironen2017sparsity}.  

\paragraph{A Sketch of Our Approach}  We propose a novel prior for $\rtau$ that goes beyond simply encouraging its value to be small, as a shrinkage prior does.  Instead, we control $\rtau$ via model-based \citep{gelman2017prior} functional regularization.  Inspired by \citet{simpson2017penalising},\footnote{We provide a detailed summary of and comparison to \citet{simpson2017penalising}'s framework in Section \ref{sec:rw}} we define a divergence function between the model of interest---denote it $p(\rvy | \rtau)$ for now---and a reference model denoted $p_{0}(\rvy)$, which does not depend on $\rtau$.  Denote the divergence as $\rkappa = \mathbb{D}[p(\rvy | \rtau) || p_{0}(\rvy)]$.  We derive $p(\rtau)$ by placing a prior on $\rkappa$ and \textit{reparameterizing} w.r.t.~$\rtau$: $\rkappa \sim \pi(\rkappa)$, $\rtau = \mathbb{D}^{-1}(\rkappa)$ where $\mathbb{D}^{-1}$ denotes the inverse of the aforementioned divergence function.  The $\rtau$-prior's density function can then be written using the change of variables formula: \begin{equation}\label{eq:priorSketch}\begin{split}
    p(\rtau) &= \pi(\rkappa) \ \left| \frac{\partial \rkappa}{\partial \rtau} \right| \\ &= \pi\left(\mathbb{D}[p(\rvy | \rtau) || p_{0}(\rvy)]\right)  \ \left| \frac{\partial \ \mathbb{D}[p(\rvy | \rtau) || p_{0}(\rvy)]}{\partial \rtau} \right|
\end{split}
\end{equation} where $|\partial \rkappa / \partial \rtau |$ is the absolute value of the divergence function's derivative w.r.t.~$\rtau$.  Note that $\mathbb{D}:\rtau \mapsto \rkappa$ must be differentiable and bijective for $p(\rtau)$ to be proper (i.e.~integrate to one).\footnote{We discuss bijectivity conditions for NNs in Section \ref{sec:apps}.}  For these conditions to be satisfied, $\rtau$ must be a scalar since $\rkappa$ is a scalar.  

These technical conditions aside, the crucial property of our framework is that the divergence is computed by integrating over $\rvy$, the random variable that corresponds to data.  In turn, the prior can represent data-space intuitions (via $\pi(\rkappa)$) and automatically translate them (via $\mathbb{D}$) into a prior on the model parameters.  This direction runs counter to how priors are usually defined: by specifying $p(\rtau)$ directly and having little-to-no information about the induced distribution on $\rvy$.  Despite its dependence on the observation model, we emphasize that our prior is not an empirical Bayesian prior \citep{casella1985introduction} since $\rvy$ is integrated out.

\section{IDEALIZED SETTING}\label{sec:ideal} 
In order to implement the prior sketched in Equation \ref{eq:priorSketch}, we need to make both theoretical choices (e.g.~the divergence) and practical choices.  We separate the description of our prior into two sections (\ref{sec:ideal} and \ref{sec:predCP}) so as to delineate theory from practice.  In this section, we provide a full theoretical description.  Section \ref{sec:predCP} then introduces some modifications that increase the scalability and broaden the applicability of our prior. 


\begin{figure*}
\centering
\subfigure[KLD Priors]{
\includegraphics[width=0.2375\linewidth]{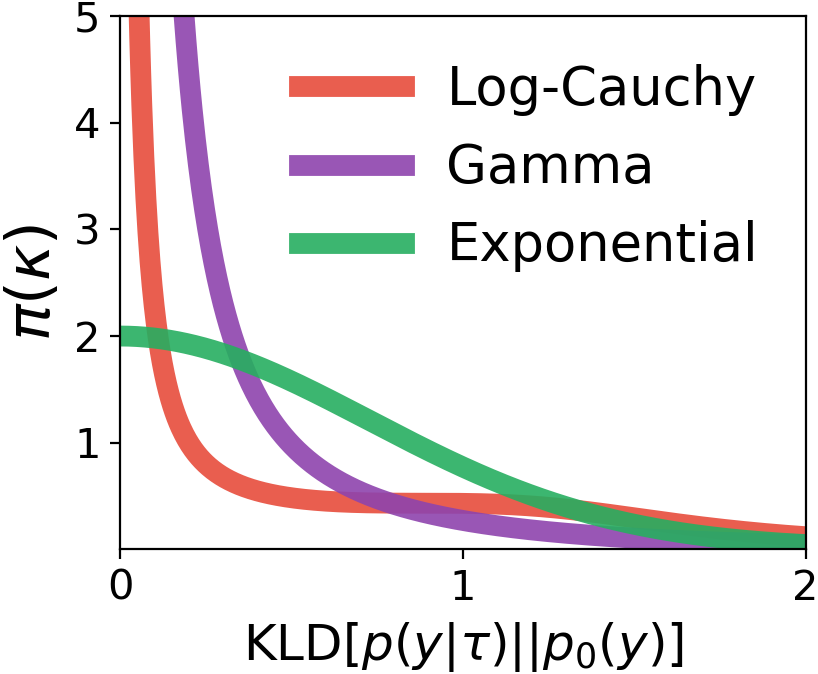}\label{subfig:pi_KLD}}
\subfigure[ECPs, $x=1$]{
\includegraphics[width=0.2375\linewidth]{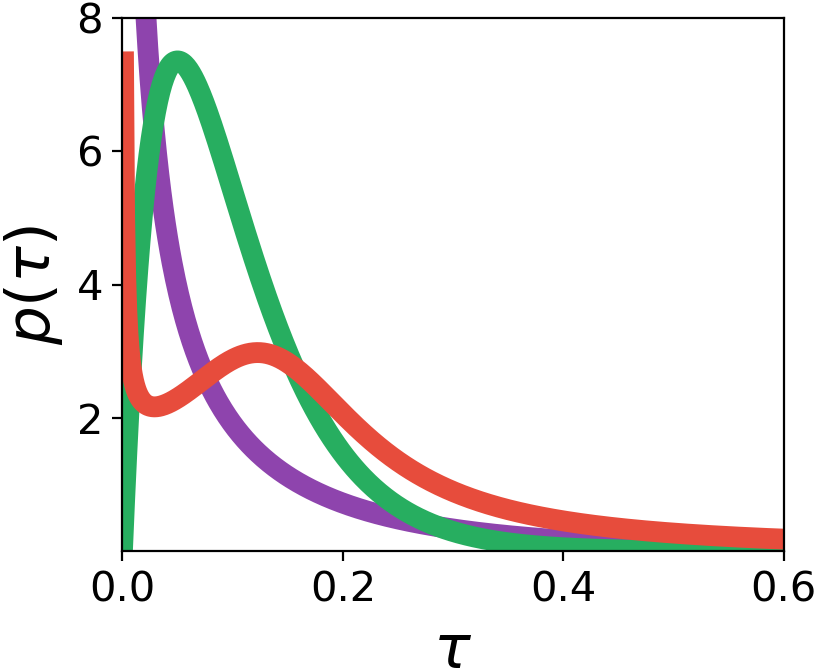}\label{subfig:lr_tau}}
\subfigure[Marginals, $x=1$]{
\includegraphics[width=0.2375\linewidth]{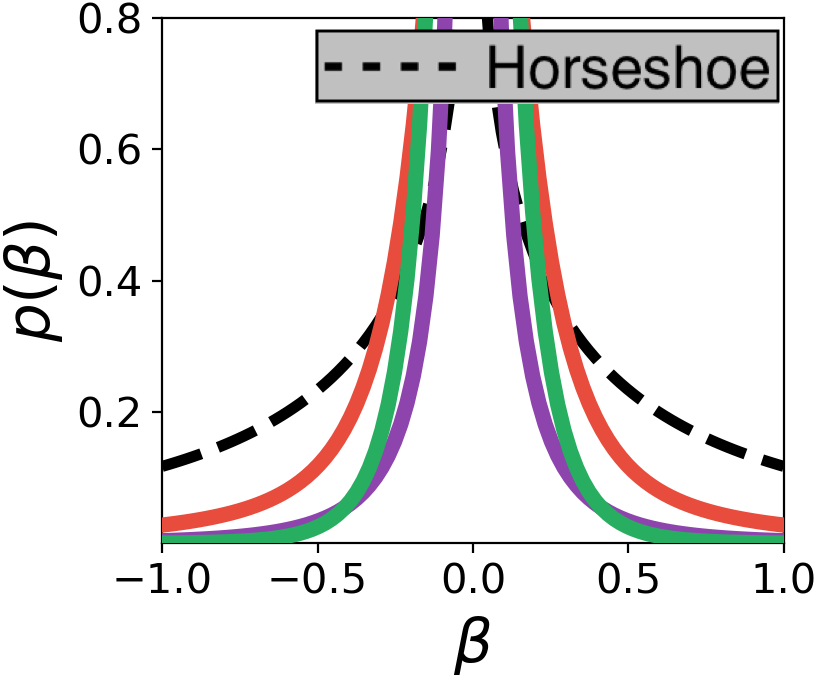}\label{subfig:lr_beta}}
\subfigure[Marginals, $x=0.25$]{
\includegraphics[width=0.2375\linewidth]{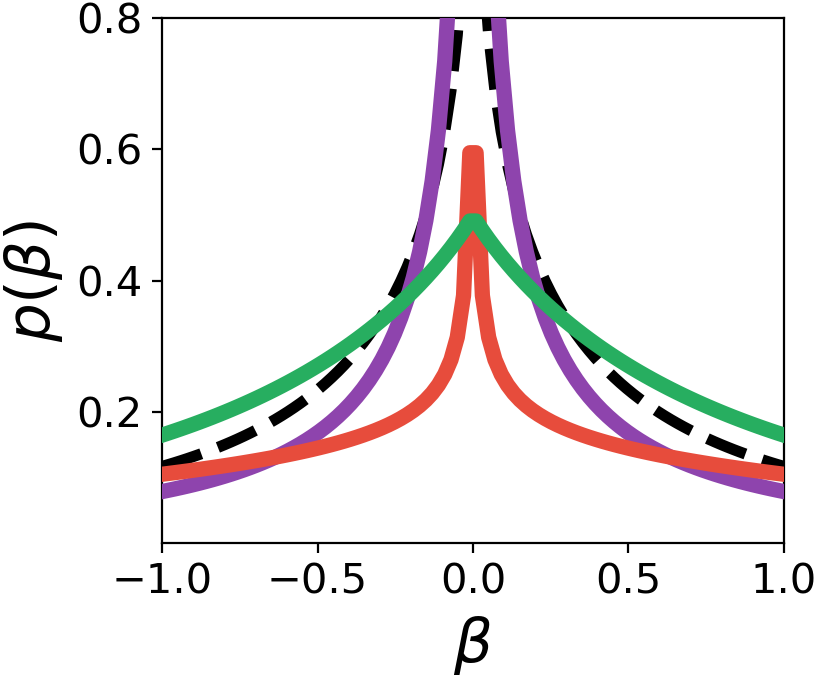}\label{subfig:lr_beta_smallX}}
\caption{\textit{ECP for Linear Regression}.  Subfigure (a) shows each KLD prior: exponential ($\lambda = .5$), gamma ($\lambda = (.2, 2)$), and log-Cauchy ($\lambda = 1$).  Subfigure (b) shows the corresponding ECP on $\rtau$.  Subfigure (c) shows the marginal prior on $\rbeta$ induced by each ECP from (b).  Subfigure (c) shows the same marginals for $x=0.25$.  The horseshoe prior \citep{carvalho2009handling} (black dashed line) is shown for reference.  The ECP adapts with the input feature, resulting in dynamic shrinkage properties.
}
\label{fig:ppcp_lrShrink}
\end{figure*}

We call the idealized version the \textit{evidence complexity prior} (ECP).  It implements Equation \ref{eq:priorSketch} via the following three steps:
\paragraph{Step \#1: Define Reference Model}  Given the model of interest in Equation \ref{eq:model}, the first step is defining the reference model $p_{0}(\rvy)$.  Crucially, $\rvy$ denotes a \emph{random variable} that corresponds to data, not an observation.  Thus, we call $p_{0}(\rvy)$ and $p(\rvy | \tau)$ `evidence functions' \citep{bishop2006pattern}, not marginal likelihoods.  In general, we define $p_{0}(\rvy)$ by replacing the first-level prior $p(\rvtheta | \rtau)$ with a less expressive prior $p_{0}(\rvtheta)$.  We obtain the final version of the reference model by marginalizing over $p_{0}(\rvtheta)$: \begin{equation}\label{eq:step_1}
    p_{0}(\rvy) = \int_{\rvtheta} \ p(\rvy | \rvtheta) \ p_{0}(\rvtheta)  \ d \rvtheta.
\end{equation}  We will exclusively use a point-mass prior $p_{0}(\rvtheta) = \delta(|\rvtheta - \vtheta_{0}|)$ so that this integration is made trivial: $p_{0}(\rvy) = p(\rvy | \vtheta_{0})$ .  To form a corresponding representation of our model of interest, we marginalize over $p(\rvtheta | \rtau)$:
\begin{equation}
   p(\rvy | \rtau) = \int_{\rvtheta} p(\rvy | \rvtheta) \ p(\rvtheta | \rtau) \  d \rvtheta.
\end{equation}  For many models, it will be hard to perform the above integration, which is why we refer to the ECP's construction as `idealized.'

\paragraph{Step \#2: Define Divergence} We next choose the divergence function.  Following \citet{simpson2017penalising}, we use the Kullback–Leibler divergence (KLD): \begin{equation}
 \text{KL}\left[ p(\rvy | \rtau) \ || \  p_{0}(\rvy) \right] = \int_{\rvy} p(\rvy | \rtau) \ \log \frac{p(\rvy | \rtau)}{p_{0}(\rvy)} \ d \rvy. 
\end{equation}  We choose this particular KLD---using $p(\rvy | \rtau)$ as the first argument---because it represents the bits lost when we approximate $p(\rvy | \rtau)$ with the reference model.  Switching the arguments would not be sensible since $\text{KL}\left[ p_{0}(\rvy) \ || \  p(\rvy | \rtau) \right]$ quantifies the bits lost when $p(\rvy | \rtau)$ approximates $p_{0}(\rvy)$, which should be easy to do since $p(\rvy | \rtau)$ is more expressive.  Other choices of divergence (e.g.~Hellinger) are possible, but in Section \ref{sec:predCP}, we will require that the divergence be a convex function w.r.t.~$p(\rvy | \rtau)$.   


\paragraph{Step \#3: Reparameterize}  Lastly, we place a prior on the divergence: $\pi(\rkappa) = \pi(\text{KL}\left[ p(\rvy | \rtau) \ || \  p_{0}(\rvy) \right])$.  The support of $\pi(\rkappa)$ should be $\mathbb{R}^{\ge 0}$ to match the KLD's codomain.\footnote{If $\pi(\rkappa)$'s support is $\mathbb{R}^{+}$, then we assume a small constant is added to the KLD so that it never evaluates to zero.}  The degree to which $\pi(\rkappa)$ favors $\rkappa=0$ represents our preference for the simpler reference model.  As $\pi(\rkappa)$ allocates more density away from zero, the functional regularization is relaxed.  Performing the change of variables then yields the ECP on $\rtau$: \begin{equation}\begin{split}
    p\left(\rtau \right) =  \pi&\left(\text{KL}\left[ p(\rvy | \rtau) \ || \  p_{0}(\rvy) \right]  \right) \\ & \ \ \ \ \ \ \ \ \ \ \ \ \times \left| \frac{\partial  \ \text{KL}\left[ p(\rvy | \rtau) \ || \  p_{0}(\rvy) \right]}{\partial \ \rtau} \right|.
\end{split}
\end{equation}
 
The crucial characteristics of the ECP are that it compares the models holistically and in data space.  Using the evidence functions (step \#1) allows the ECP to directly assess how $\rtau$ affects $\rvy$.  Hence the ECP embraces the philosophy that the prior can only be understood in the context of the likelihood \citep{gelman2017prior}.  Computing the KLD (step \#2) then provides a functional comparison of how the models allocate probability in data space.  This data-space behavior is our ultimate concern for black-box models.  

\subsection{Example: Linear Regression}  We continue discussion of the ECP with a concrete example.  Consider the linear model $ \mathbb{E}[\ry | x, \rbeta] = x \rbeta$, where $\ry \in \mathbb{R}$ denotes a scalar response, $x \in \mathbb{R}$ its covariate / feature, and $\rbeta \in \mathbb{R}$ the model parameter.  While this is undoubtedly a simple example, the priors we later describe for NNs will have commonalities.  Let us now step through the ECP derivation.  We choose the first-level prior to be normal, $p(\rbeta | \rtau) = \text{N}(0, \rtau)$, and the reference prior\footnote{We use `reference prior' to refer to the prior for the reference model, not to \citet{bernardo1979reference}'s class of objective priors.} to be `the spike,' $p_{0}(\rbeta) = \delta(|\rbeta - 0|)$.  The ECP for $\tau$ is then: \begin{equation}\begin{split}
    p(\rtau; x) =  &\pi\left( \text{KL}\left[ \text{N}(0, \sigma_{y}^{2} + x^{2} \rtau) \ || \  \text{N}(0, \sigma_{y}^{2}) \right] \right) \\ &  \times \left| \frac{\partial  \ \text{KL}\left[ \text{N}( 0, \sigma_{y}^{2} + x^{2} \rtau) \ || \  \text{N}(0, \sigma_{y}^{2}) \right]}{\partial \ \rtau} \right|
\end{split}
\end{equation} where $\sigma^{2}_{y}$ is the response noise.  In this case---and for conditional models in general---the ECP is a function of the features $x$ and any other independent variables.  Other default priors such as the \textit{$g$-prior} \citep{zellner1986gPrior} and Jeffreys prior \citep{jeffreys1946invariant} have this dependence as well, which reflects their holistic natures.

\paragraph{Choosing $\pi(\rkappa)$}  We next discuss the choice of $\pi(\rkappa)$ and its effect on the resulting ECP.  Figure \ref{subfig:lr_tau} shows three ECPs, each defined by a different choice of KLD prior (\ref{subfig:pi_KLD}): exponential (green), gamma (purple), and log-Cauchy (red).  The choice of $\pi(\rkappa)$ is significant.  First considering the exponential prior, it clearly favors $\rtau > 0$ since the density function decays to zero at the origin.  We can interpret this behavior in the context of the reference and original models as a strict preference for $p(\rvy | \rtau)$.  At the other extreme is the gamma prior: it has a mode at $\rtau=0$ and then quickly decays as $\rtau$ increases.  Thus, the gamma strictly prefers $p_{0}(\rvy)$.  Last we have the log-Cauchy, which we chose due to its heavy tail.  Heavy-tailed priors have been well-validated for robust regression since they allow the shrinkage to be ignored under sufficient counter-evidence \citep{carvalho2009handling}.  A similar logic can be applied to the KLD: perhaps the reference model is too simplistic and $p(\rvy | \rtau)$ is drastically superior.  If so, we want any preference for the reference model to be forgotten.  Figure \ref{subfig:lr_tau} shows that the log-Cauchy results in an ECP with two modes, one at $\rtau=0$ and another at $\rtau \approx .15$.  The log-Cauchy is able to balance its preferences for $p(\rvy | \rtau)$ and $p_{0}(\rvy)$, interpolating between the exponential and gamma's single-mindedness.

\paragraph{Marginal Priors and Feature Dependence}  It is perhaps more intuitive to examine the marginal prior on $\rbeta$ induced by the ECP: $p(\rbeta) = \int p(\rbeta | \rtau) p(\rtau) d\rtau$.  Figure \ref{subfig:lr_beta} shows the marginal prior for the three ECPs considered above and compares them to the horseshoe prior \citep{carvalho2009handling} (black dashed line).  The three priors behave as expected from looking at $p(\rtau)$: the gamma shrinks the hardest and the log-Cauchy has the heaviest tails.  Yet, recall that the ECP also depends on $x$.  So far we have assumed $x=1$, but in Figure \ref{subfig:lr_beta_smallX} we show the same marginal priors for $x=0.25$.  This change in $x$ results in drastically different ECPs.  As $x \rightarrow 0$, the ECP (no matter the choice of $\pi(\rkappa)$) becomes heavier tailed, allowing more deviation from the reference model.  This behavior is natural since, when $x$ is small, large $\rbeta$ values are necessary to substantially change the model's predictions.  See the appendix for more discussion, including the ECP for multivariate regression.

\section{PREDICTIVE COMPLEXITY PRIORS}\label{sec:predCP}
We now move on to our primary contribution: deriving a prior that has the same holistic, function-space properties as the ECP but is tractable for models such as NNs.  As mentioned earlier, the primary weakness of the ECP is the difficulty of step \#1: integrating over $\rvtheta$.  In this section, we propose modifications to the ECP derivation that result in a tractable and scalable alternative.  We call the resulting prior a \textit{predictive complexity prior} (PredCP).


\paragraph{KLD Upper Bound}  As the primary obstacle is integrating over $p(\rvtheta | \rtau)$, we make headway by defining the PredCP using the following upper bound on the ECP's KLD:
\begin{equation}\label{eq:KLD_upper}\begin{split}
\text{KL}[ p(\rvy | \rtau)  || p_{0}(\rvy) ] &= \text{KL}\left[ \mathbb{E}_{\rvtheta | \rtau} \left[ p(\rvy | \rvtheta) \right] ||  p_{0}(\rvy) \right] \\ &\le  \mathbb{E}_{\rvtheta | \rtau}   \text{KL}\left[ p(\rvy | \rvtheta) ||  p_{0}(\rvy) \right]. 
\end{split}\end{equation}  We arrive at the upper bound via the strict convexity of $\text{KL}\left[ p(\rvy | \rvtheta) \ || \  p_{0}(\rvy) \right]$ and Jensen's inequality.  The bound reverses the order in which marginalization and divergence computation are done for the ECP.  This reversal makes the PredCP more practical since its KLD is taken between the data models.  These are usually simple distributions (e.g.~categorical, Gaussian) that afford a closed-form KLD.  Unfortunately, the expectation over $\rvtheta | \rtau$ may still not be analytically available.  We recommend evaluating the integral using a \textit{differentiable, non-centered} Monte Carlo (MC) approximation \citep{kingma2014efficient}.  Doing so ensures the KLD's derivative w.r.t.~$\rtau$ is well-defined.  For supervised learning, a downside of the upper bound is that the dependencies between predictions are lost.  Having the data model factorize across feature observations---$p(\rmY | \mX, \rvtheta) = \prod_{n} p(\rvy | \vx_{n}, \rvtheta)$---results in the KLD becoming a point-wise sum.  This is not an issue for the unsupervised case since there is no concept of features.



\paragraph{Mini-Batching}  For supervised learning, evaluating the PredCP requires a sum over all feature observations, which will be computationally costly for large data sets.  Therefore we recommend the PredCP be evaluated with mini-batches.  Moreover, we compute the KLD's mean across the mini-batch, not the sum.  In doing so we assume that the batch's mean KLD represents an unbiased estimate of the full-data mean KLD.  We use the mean KLD primarily for practical purposes: it is easier to set $\pi(\rkappa)$'s parameters since they do not have to account for the batch size.

\begin{figure}
\centering
\includegraphics[width=0.6\columnwidth]{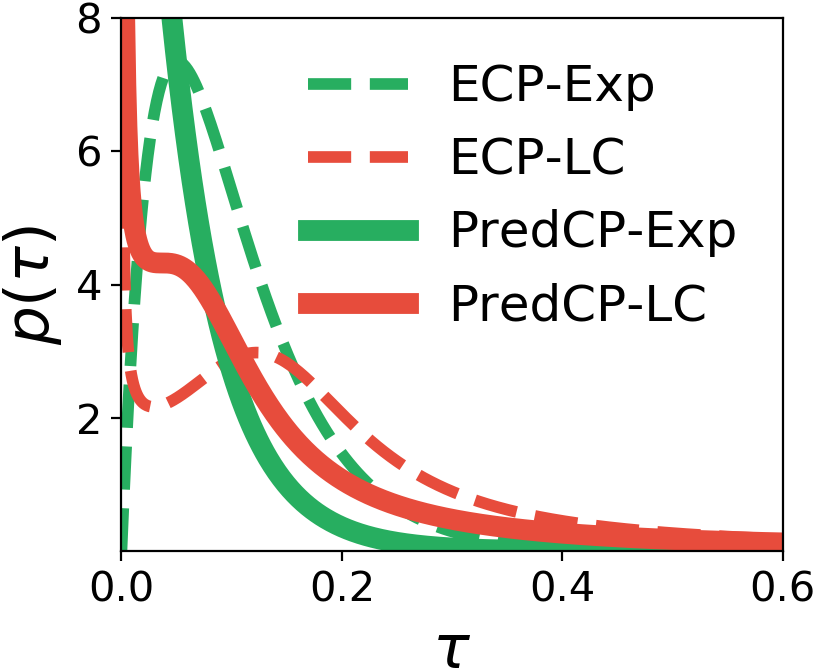}
\caption{\textit{PredCP for Linear Regression}.  The ECP (dashed lines) vs the PredCP (solid lines) for the exponential and log-Cauchy KLD priors ($x=1$).
}
\label{fig:ppcp_lrShrink_PredCP}
\end{figure}

\paragraph{PredCP Final Form}
Below we give the final form of the PredCP for supervised learning, combining the point-mass reference prior, the KLD upper bound, and mini-batching: \begin{equation}\begin{split}
    p(\rtau; & \mX_{B}) = \left| \frac{1}{B} \sum_{b=1}^{B}  \frac{\partial  \ \mathbb{E}_{\rvtheta | \rtau}\text{KL}_{b}}{\partial \ \rtau} \right| \ \times \\ &\pi\left( \frac{1}{B} \sum_{b=1}^{B} \mathbb{E}_{\rvtheta | \rtau} \text{KL}\left[ p(\rvy | \vx_{b}, \rvtheta) \ || \  p(\rvy | \vx_{b}, \vtheta_{0}) \right]  \right)
\end{split}
\end{equation} where $b$ indexes the $B$-sized batch and $\mathbb{E}_{\rvtheta | \rtau}\text{KL}_{b}$ is shorthand for the expected KLD for the $b$th instance.  The PredCP encourages stronger shrinkage than the ECP, which is expected due to the upper bound.  For a given $\rtau$, the PredCP deems the models to be more discrepant than the ECP would for the same $\rtau$.  This is an appropriate inductive bias for the PredCP since it will be used for large models that often require strong regularization.  In Figure \ref{fig:ppcp_lrShrink_PredCP}, we compare the ECP (dashed lines) and the corresponding PredCP (solid lines) for the linear regression example.  The PredCP's inductive bias is evident in the leftward shift of the density functions.  This shift can change the PredCP's behavior considerably in comparison to the corresponding ECP.  The exponential's PredCP has a mode at $\rtau=0$ whereas its ECP decays to zero at the origin. 

\section{APPLICATIONS OF THE PREDCP}\label{sec:apps}
We now demonstrate the PredCP's utility for modern machine learning.  We consider two applications: depth-selection for Bayesian NNs \citep{dikov2019bayesian, nalisnick19dropout, antoran2020variational} and sharing statistical strength across tasks for meta-learning \citep{chen2019modular}.  Both of these applications exhibit the PredCP's ability to enable Bayesian reasoning across the model's macro-structures (e.g.~layers) while still being a tractable and proper prior.

\paragraph{Bijectivity Conditions for Neural Networks}  Before moving on to these two applications, we first address some technical conditions.  Recall that for the PredCP to be a proper prior (i.e.~integrate to one), the expected KLD must be differentiable and bijective w.r.t.~$\rtau$.  The former is easy to satisfy by using a non-centered MC approximation, as mentioned above.  It is not obvious if the latter is satisfied by NNs.  One could check the condition via brute force, by using numerical integration.  Since $\rtau$ is a scalar, the numerical solution should be stable and makes for a good unit test.  Yet, we show in the appendix that bijectivity is satisfied for feedforward NNs with ReLU activations and Gaussian or categorical observation models.  No architectural modifications are necessary.

\begin{figure*}
\centering
\subfigure[Joint Density Function]{
\includegraphics[width=0.46\linewidth]{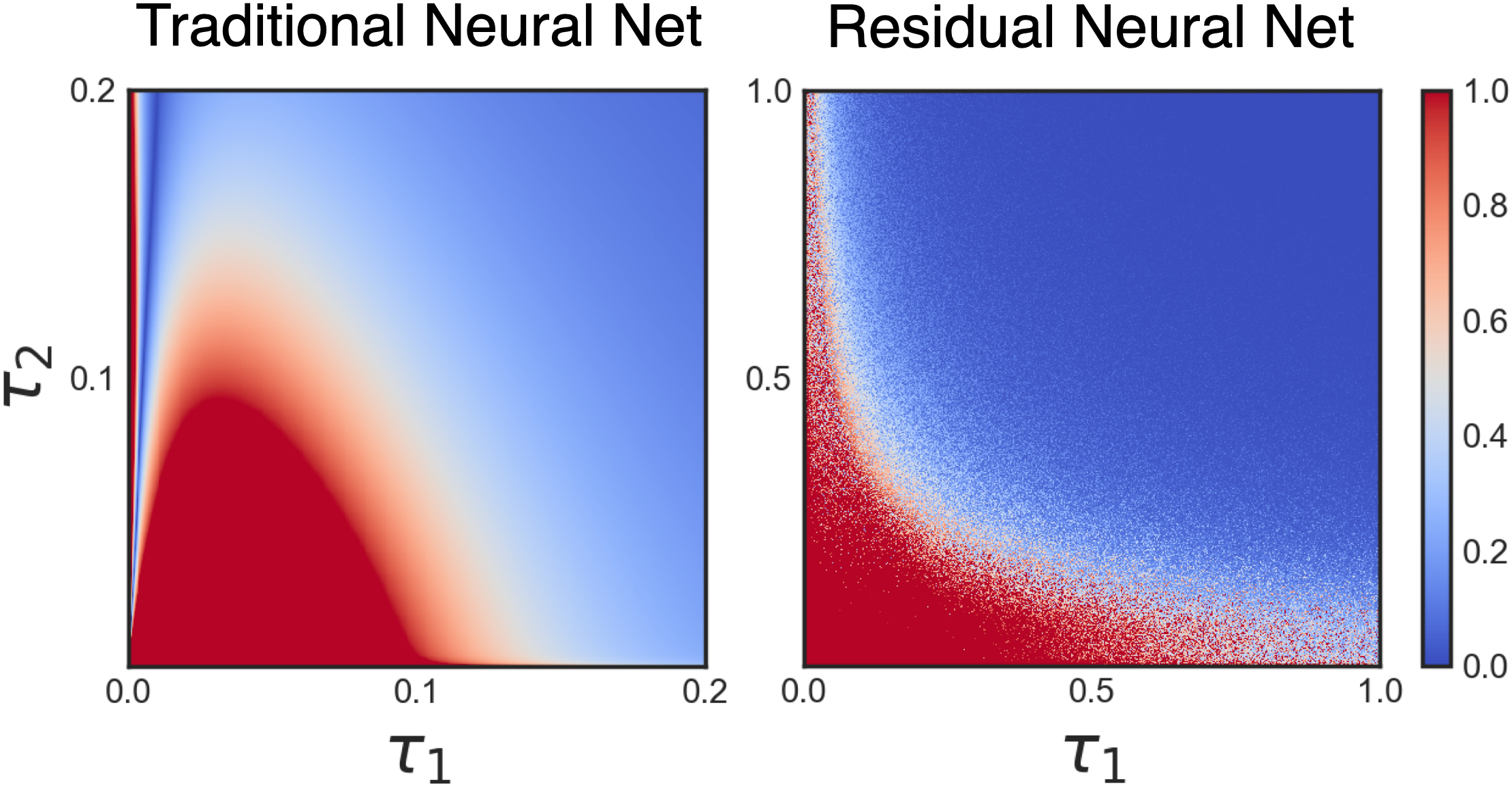}\label{subfig:joint}}
\hfill 
\subfigure[Functions Sampled from Resnet]{
\includegraphics[width=0.52\linewidth]{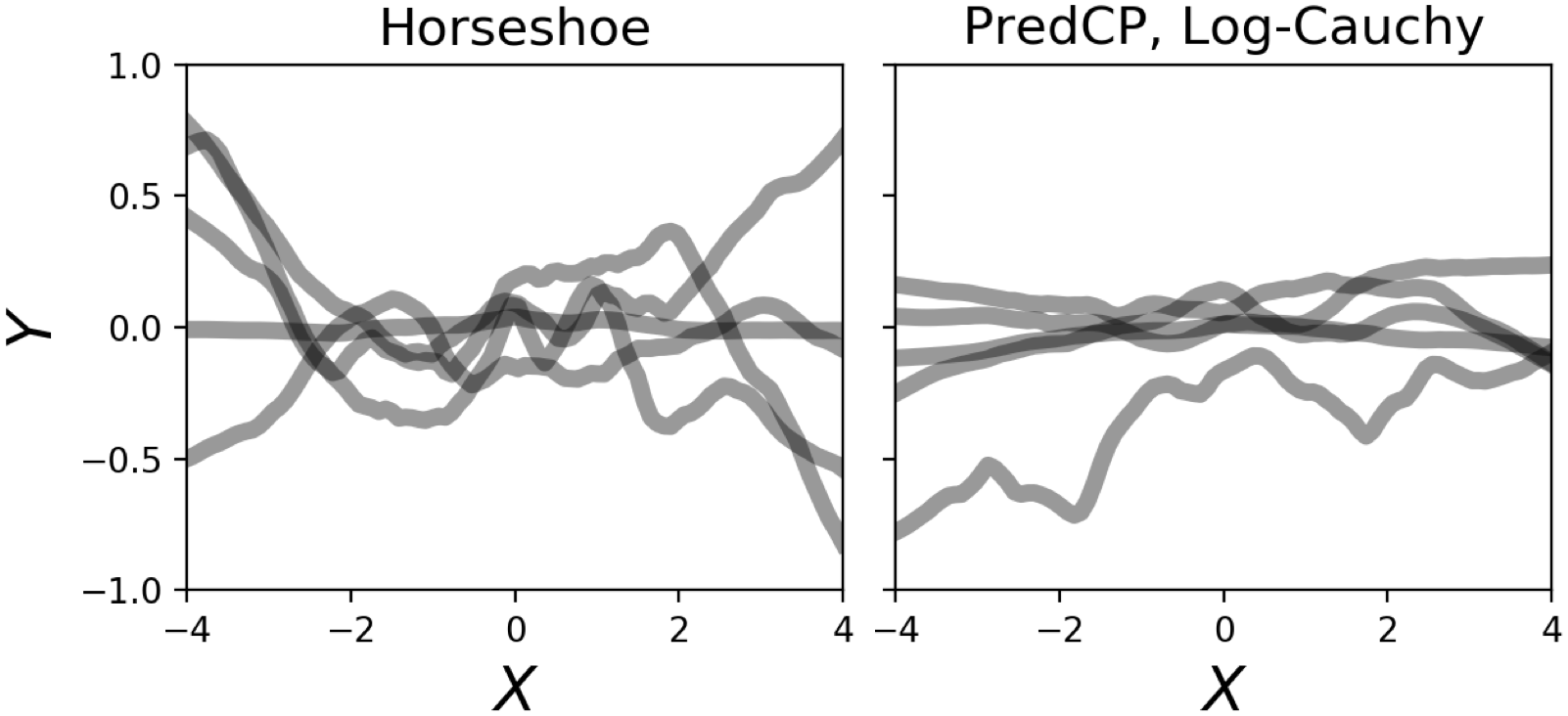}\label{subfig:f_samples}}
\caption{\textit{Depth-Wise PredCP}.  Subfigure (a) shows $\pi(\rtau_{1}, \rtau_{2})$ for traditional (left) and residual (right) NNs.  Subfigure (b) shows functions sampled from a resnet with a PredCP and a horseshoe prior for comparison.  The KLD prior is Log-Cauchy$(0,1)$ in both cases.
}
\label{fig:ppcp_NN}
\end{figure*}

\subsection{Depth Selection for Neural Networks}  PredCPs allow us to perform Bayesian reasoning over the depth of a NN.  First assume the NN to be a residual network (resnet) \citep{he2016Deep}; later we will address the traditional feedforward case.  Since we wish to isolate the effect of depth, we choose the reference model to have one fewer layer ($l-1$ layers) than the model of interest ($l$ layers).  The KLD between these models will then capture the extra capacity afforded by the additional layer.  

More formally, for an arbitrary layer $l$, the prior on the (square) weight matrix $\rmW_{l} \in \mathbb{R}^{D_{h} \times D_{h}}$ for the reference and original models are: $p_{0}(\rmW_{l}) = \delta\left( | \rmW_{l} - \mathbf{0} | \right)$, $p(\rmW_{l} | \rtau_{l}) = \text{N}(\mathbf{0}, \ \rtau_{l} \mSigma_{l})$ where $\rtau_{l}$ is again the parameter of interest.  Integrating over $p_{0}$ sets $\rmW_{k} = \mathbf{0}$ for $k \ge l$ for the reference model and $k > l$ for the original model.  The resnet then maps the hidden layers directly to the output layer, thereby allowing the PredCP to compare the predictions when using $l-1$ vs $l$ layers.  We can define the PredCP for all layers by applying the above priors recursively from the bottom-up: 
\begin{equation}\label{depthPREDCP}\begin{split}
      p&(\rtau_{1}, \ldots, \rtau_{L}) = p(\rtau_{1}) \prod_{l=2}^{L} p(\rtau_{l} | \rtau_{1}, \ldots, \rtau_{l-1}) \\ 
    &= \prod_{l=1}^{L} \pi\left( \mathbb{D}(\rtau_{l}; \rtau_{1:l-1} )  \right) \ \left| \frac{\partial  \ \mathbb{D}(\rtau_{l}; \rtau_{1:l-1}) }{\partial \ \rtau_{l}} \right|
\end{split}\end{equation} where the divergence function is
\begin{equation*}\begin{split}
       &\mathbb{D}(\rtau_{l}; \rtau_{1:l-1} ) = \\ &\mathbb{E}_{\{\rmW_{j} | \rtau_{j}\}_{j=1}^{l}}  \text{KL}\left[ p(\rvy | \{\rmW_{j} \}_{j=1}^{l}) \ || \  p(\rvy | \{\rmW_{k} \}_{k=1}^{l-1}) \right]. 
\end{split}\end{equation*}  

Computing the full prior requires $L$ forward propagations, each evaluating a progressively deeper network with the hidden units at layer $l$ serving as the last hidden layer.  In practice, we cache the forward propagation required to evaluate the original model for $\rtau_{l}$ and use it as the reference model when evaluating the prior for $\rtau_{l+1}$.  Nearly the same procedure can be applied to non-residual networks, except that the residual connections can no longer be relied upon to transport the hidden units to the output layer.  Rather, the network must be `short circuited,' with the final hidden units being directly multiplied with the output weights.  

Figure \ref{subfig:joint} shows the joint density function $\pi(\rtau_{1}, \rtau_{2})$ for both traditional and residual NNs.  The PredCP's capability for depth selection is conspicuous for the traditional NN (left).  The high density region (red) touches the $x$-axis but not the $y$-axis except near the origin.  This implies that $\rtau_{2}$ cannot grow unless $\rtau_{1} > 0$, meaning that the first layer is activated.  For resnets (right), the density's \textit{L}-shape means that either layer can be active while the other is inactive ($\rtau \approx 0$), which is made possible by the skip connection.  Yet the density's bias towards the $x$-axis suggests that the resnet-PredCP still prefers to activate $\rtau_{1}$'s layer before activating $\rtau_{2}$'s.

Further intuition can be had by examining the depth-wise PredCP for resnets with a Gaussian data model.  Denote a hidden layer for the $n$th observation as $\rh_{n,l} = \rh_{n,l-1} + f_{l}(\rh_{n,l-1} \rmW_{l})$ and the output weights as $\rmW_{o} \sim \text{N}(\mathbf{0},\mSigma)$.  We assume $\rmW_{l}$ is parameterized as $\sqrt{\rtau}_{l} \tilde{\rmW}_{l}$, $\tilde{\rmW}_{l} \sim \text{N}(\mathbf{0}, \mSigma)$.  The expected KLD for computing $\pi(\rtau_{l} | \rtau_{1:l-1})$ (Equation \ref{depthPREDCP}) is then:
\begin{equation}\begin{split}
    &\mathbb{E}_{\{\rmW_{j} | \rtau_{j}\}_{j=1}^{l}}  \text{KL}\left[ p(\rvy | \{\rmW_{j} \}_{j=1}^{l}) ||  p(\rvy | \{\rmW_{k} \}_{k=1}^{l-1}) \right] \\ & = \frac{\rtau_{l}}{2\sigma^{2}_{y}} \frac{1}{N} \sum_{n=1}^{N} \text{Var}_{\tilde{\rmW}, \rmW_{o}} \left[f_{l}(\rh_{n,l-1} \tilde{\rmW}_{l})\rmW_{o}\right] 
\end{split}
\end{equation} where $\sigma^{2}_{y}$ denotes the response noise and $f_{l}$ is any positively homogeneous activation function (such as the ReLU).  The crucial term $\text{Var}[\rvf_{l} \rmW_{o}]$ represents the variance that the $l$th layer's transformation term contributes to the resnet's prediction for $\vx_{n}$.  The expression makes clear that the PredCP is performing functional regularization: a zero-favoring $\pi(\rkappa)$ will encourage this variance to be small.  Dropout has been shown to curb the variance of hidden units in a similar way \citep{baldi2013understanding}.  Figure \ref{subfig:f_samples} shows functions sampled from a resnet with a horseshoe prior (left) and a depth-wise log-Cauchy PredCP (right).  The PredCP's samples are closer to linear due to the regularization of the predictive variance.  Yet, recall that the log-Cauchy is heavy-tailed and therefore allows some functions to stray from the origin, as we see one sample has done.

\subsection{Hierarchical Modeling for Meta-Learning}  
Meta-learning is another natural application for the PredCP as it can control the degree to which information is pooled across tasks.  Following the approach of \citet{chen2019modular}, we use the generative model: $\mathcal{D}_{t} \sim p(\mathcal{D}_{t} | \rvtheta_{t}), \ \rvtheta_{t} \sim \text{N}(\vphi, \rtau \mathbb{I})$ where $t$ indexes the task, $\mathcal{D}_{t}$ is data for the $t$th task, $\rvtheta_{t}$ are local parameters specific to the $t$th task, and $\{\vphi, \rtau \}$ are global \textit{meta-parameters}.  The scale $\rtau$ controls local adaptation, and as $\rtau \rightarrow 0^{+}$, the task structure becomes irrelevant.  This hierarchical meta-learning model is perfectly suited for a PredPC as the global parameters $\vphi$ define a natural reference model: \begin{equation*}
    p_{0}(\mathcal{D}_{t}) = \int_{\rvtheta_{t}}  p(\mathcal{D}_{t} | \rvtheta_{t}) \ \delta(|\rvtheta_{t} - \vphi|) \ d\rvtheta_{t} = p(\mathcal{D}_{t} | \vphi).
\end{equation*}  Computing the KLD between the original and reference models then quantifies the information lost when ignoring the task structure.  The prior is written as: \begin{equation}\begin{split}
     p(\rtau) & \ = \ \left| \frac{1}{T} \sum_{t} \frac{\partial  \  \mathbb{E}_{\rvtheta_{t} | \rtau}\text{KL}_{t}}{\partial \ \rtau} \right| \ \times \\ & \pi\left( \frac{1}{T} \sum_{t=1}^{T} \mathbb{E}_{\rvtheta_{t} | \rtau}\text{KL}\left[  p_{+}(\mathcal{D}_{t} | \rvtheta_{t}) \ || \   p_{0}(\mathcal{D}_{t} | \vphi) \right]  \right)
\end{split}
\end{equation} 
where $\mathbb{E}_{\rvtheta_{t} | \rtau}\text{KL}_{t}$ is shorthand for the expected KLD on the $t$th task.  In the experiments, we follow \citet{chen2019modular}'s modular specification by applying the PredCP layer-wise.  Doing so allows the feature-extracting shallow layers to adapt to a different degree than the classification-based final layer.  

\begin{table*}
\centering
\caption{\textit{Logistic Regression}.  We report test set predictive log-likelihoods for the half-Cauchy prior, ECP, and PredCP under both VI and MCMC.  Results are averaged across $20$ splits.
}
\resizebox{\textwidth}{!}{%
\begin{tabular}{l cc | ccc | ccc }
 \multicolumn{3}{c}{} & \multicolumn{3}{c}{\textsc{Variational Inference}} & \multicolumn{3}{c}{\textsc{Markov Chain Monte Carlo}} \\
\textsc{Data Set} & \textsc{N}$_{\text{train}}$ & D & \textsc{Half-Cauchy} & \textsc{\textbf{ECP}} &  \textsc{\textbf{PredCP}} & \textsc{Half-Cauchy} & \textsc{\textbf{ECP}}  & \textsc{\textbf{PredCP}} \\ 
\midrule
\texttt{allaml} & 51 & 7129 & $\minus 0.43 \scriptstyle{ \pm .01}$ & $ \mathbf{\minus 0.32} \scriptstyle{ \pm .01}$ & $\mathbf{\minus 0.32} \scriptstyle{ \pm .01}$ &  $\minus 0.19 \scriptstyle{ \pm .02}$ & $\mathbf{\minus 0.17} \scriptstyle{ \pm .02}$ & $ \mathbf{\minus 0.17} \scriptstyle{ \pm .02}$   \\
\texttt{colon} & 44 & 2000 & $\mathbf{\minus 0.61} \scriptstyle{ \pm .02}$ & $ \minus 0.63 \scriptstyle{ \pm .03}$ & $\minus 0.66 \scriptstyle{ \pm .02}$ & $\minus 0.54 \scriptstyle{ \pm .05}$ & $ \mathbf{\minus 0.52} \scriptstyle{ \pm .05}$ & $\minus 0.54 \scriptstyle{ \pm .04}$  \\
\texttt{breast} & 82 & 9 & $ \minus 0.60 \scriptstyle{ \pm .01}$ & $\mathbf{\minus 0.58} \scriptstyle{ \pm .01}$ & $\mathbf{\minus 0.58} \scriptstyle{ \pm .01}$ & $\minus 0.55 \scriptstyle{ \pm .02}$ &  $\minus 0.55 \scriptstyle{ \pm .01}$ & $\minus 0.55 \scriptstyle{ \pm .02}$  \\
\end{tabular}%
}
\label{tab:LRresults}
\end{table*}

\section{RELATED WORK}\label{sec:rw}

\paragraph{Penalized Complexity Prior} Our work is directly inspired by and extends \citet{simpson2017penalising}'s \textit{penalized complexity prior} (PCP).  Let $p_{0}(\rvtheta)$ denote a `base' prior and $p_{+}(\rvtheta | \rtau)$ an `extended' prior, with $\rtau$ controlling $p_{+}$'s expressivity.  \citet{simpson2017penalising} define a prior for $\rtau$ by placing a prior on the root-KLD and changing variables:  
 \begin{equation*}\begin{split}
p&\left(\rtau \right) = \pi\left(\sqrt{2\text{KL}\left[ p_{+}(\rvtheta | \rtau) \ || \  p_{0}(\rvtheta) \right]} \right) \ \left| \frac{\partial  \ \sqrt{2\text{KL}}}{\partial \ \rtau} \right|.
 \end{split}\end{equation*}  The major difference between the PCP and our PredCP is in how the KLD is formulated.  \citet{simpson2017penalising} compute the divergence \emph{between priors} $p(\rvtheta)$ whereas we use the divergence \emph{between data models} $p(\rvy | \rvtheta)$.  This crucial change is necessary since the PCP is hard to define for NNs and similarly complicated models.  As the priors \emph{and} the divergence are defined in $\rvtheta$-space, specifying the PCP still requires intimate knowledge of the parameters, running into the same challenges of high-dimensionality and unidentifiability.  Because of our modification to $\rvy$-space, this makes the computation harder since we need to marginalize $\rvtheta$ to work with $p(\rvy | \rtau)$.  This difficulty necessitated the KLD upper bound in Equation \ref{eq:KLD_upper}.  Another benefit of comparing the models in $\rvy$-space is that we can easily use point-mass priors for $\rvtheta$.  \citet{simpson2017penalising} also do this, but because their KLD is defined on $\rvtheta$, they need to take limits.  PCPs have been used as priors for P-splines \citep{ventrucci2016penalized}, distributional regression \citep{klein2016scale}, autoregressive processes \citep{sorbye2017penalised}, mixed effects models \citep{ventrucci2019pc}, and Gaussian random fields \citep{fuglstad2019constructing}.
 
\paragraph{Functional Priors for Bayesian NNs}  Our work is also motivated by recent efforts to rethink prior specification for Bayesian NNs.  As we truly care about the distribution over predictive functions, specifying \textit{functional priors} has received much attention of late \citep{ma19implicit, hafner2018noise, flam2018characterizing,louizos2019functional}.  The hope is that it is easier to reason about our preferences for functions than for parameters.  However, existing functional priors introduce cumbersome byproducts into the Bayesian workflow.  Placing a functional prior on a NN requires either taking infinite width limits \citep{pearce2019expressive}, optimizing divergences involving stochastic processes \citep{flam2017mapping, sun2018functional}, or pre-training \citep{flam2017mapping, nalisnick2018learning, atanov2018the}.  Our framework, on the other hand, uses reparameterization to obtain a proper prior on the parameters, creating no complications for traditional Bayesian inference.

\begin{table*}
\centering
\caption{\textit{ARD-ADD Resnet}.  We report test set RMSE for UCI benchmarks, comparing the PredCP against a shrinkage prior \citep{nalisnick19dropout} and a fixed scale.  Results are averaged across $20$ splits.
}
\resizebox{.95\textwidth}{!}{%
\begin{tabular}{l c c  c c c c c }
\\
 Prior Type & \texttt{boston} & \texttt{concrete} &  \texttt{energy} &  \texttt{kin8nm} & \texttt{power} & \texttt{wine} & \texttt{yacht}\\ 
\midrule
\textsc{Fixed}& $ 2.29 \ \scriptstyle{\pm .33}$ & $\mathbf{3.51} \ \scriptstyle{\pm .41}$ & $0.83 \ \scriptstyle{\pm .14}$ & $0.06 \ \scriptstyle{\pm .00}$ & $3.32 \ \scriptstyle{\pm .09}$ & $0.58 \ \scriptstyle{\pm .04}$ & $0.66 \ \scriptstyle{\pm .12}$ \\
\textsc{Shrinkage} & $2.37 \ \scriptstyle{\pm .18}$ & $3.76 \ \scriptstyle{\pm .23}$ &  $0.85 \ \scriptstyle{\pm .08}$ & $0.06 \ \scriptstyle{\pm .00}$ &  $\mathbf{3.24} \ \scriptstyle{\pm .07}$ & $\mathbf{0.54} \ \scriptstyle{\pm .03}$ & $0.60 \ \scriptstyle{\pm .16}$\\
\textsc{\textbf{PredCP}}& $ \mathbf{2.26} \ \scriptstyle{\pm .06}$ & $3.70 \ \scriptstyle{\pm .46}$ & $\mathbf{0.82} \ \scriptstyle{\pm .07}$ & $0.06 \ \scriptstyle{\pm .00}$ & $3.27 \ \scriptstyle{\pm .09}$  & $0.56 \ \scriptstyle{\pm .03}$ & $\mathbf{0.57} \ \scriptstyle{\pm .03}$ \\ 
\end{tabular}%
}
\label{tab:UCIresults}
\end{table*}

\begin{table*}[h]
\centering
\caption{\textit{Few-Shot Learning}.  We report test set accuracy for the PredCP, comparing it to a shrinkage prior, a uniform prior \citep{chen2019modular}, and non-Bayesian MAML.
}
\resizebox{.8\textwidth}{!}{%
\begin{tabular}{l  cc | cc }
& \multicolumn{2}{c}{\textsc{Fewshot-CIFAR100}} & \multicolumn{2}{c}{\textsc{Mini-ImageNet}} \\
 & \textsc{1-Shot} &  \textsc{5-Shot} & \textsc{1-Shot} &  \textsc{5-Shot} \\
\midrule
MAML & $35.6 \pm 1.8$ & $50.3 \pm 0.9$  & $46.8 \pm 1.9$ & $58.4 \pm 0.9$   \\
$\sigma$-MAML + uniform prior & $ 39.3 \pm 1.8$  & $51.0 \pm 1.0$ & $47.7 \pm 0.7$ & $60.1 \pm 0.8$  \\
$\sigma$-MAML + shrinkage prior & $40.9 \pm 1.9$   &  $52.7 \pm 0.9$  & $48.5 \pm 1.9$ & $60.9 \pm 0.7$   \\
$\sigma$-MAML + \textbf{PredCP} & $\textbf{41.2} \pm 1.8$ & $\textbf{52.9} \pm 0.9$  & $\mathbf{49.3} \pm 1.8$ & $\mathbf{61.9} \pm 0.9$   \\
\end{tabular}%
}
\label{tab:FSresults}
\end{table*}

\section{EXPERIMENTS}
We evaluate the PredCP on regression, classification, and few-shot learning tasks under a variety of algorithms for posterior inference.  The experimental details are provided in the appendix.  In all cases, we use relatively small data sets so that the prior's influence is not overwhelmed by the likelihood.

\paragraph{Logistic Regression}  We first report an experiment in which the ECP can be computed and posteriors obtained with high-fidelity.  We use the logistic regression model: $\ry \sim \text{Bernoulli}(f(\vx \rvbeta))$, $\rbeta_{d} \sim \text{N}(0, \rlambda_{d}^{2}\rtau^{2})$, $\rlambda_{d} \sim \text{C}^{+}(0,1)$ where $f$ denotes the logistic function and $\text{C}^{+}$ a half-Cauchy prior.  We compare three priors for $\rtau$: $\text{C}^{+}(0,1)$, which is the default prior recommended by \citet{gelman2006prior} and \citet{carvalho2009handling}, the ECP (via probit approximation \citep{bishop2006pattern}), and the PredCP.  The log-Cauchy$(0,1)$ is the KLD prior.  We use \texttt{Stan} \citep{carpenter2017stan} to obtain the full posterior $p(\rvbeta, \rvlambda, \rtau | \mX, \vy)$, performing both variational inference (Normal mean-field approximation) \citep{kucukelbir2017automatic} and Hamiltonian MC.  We test the priors on three small medical data sets \citep{golub1999molecular, alon1999broad, patricio2018using} so that the prior strongly influences the posterior.  Furthermore, two of the data sets are high-dimensional ($2000+$) in order to test if the PredCP can prevent overfitting.  Table \ref{tab:LRresults} reports the predictive log-likelihood on the test set averaged over $20$ splits.  The ECP and PredCP have comparable performance and outperform the half-Cauchy in four of six cases and with one tie.

\paragraph{Neural Networks}  We next report results using resnets for regression: $\ry \sim \text{N}(\ry | \vx, \{ \rmW_{l}\}_{l=1}^{3})$, $\rw_{l, i, j} \sim \text{N}(0, \rlambda_{l,i}^{2}\rtau_{l})$, $\rlambda_{l,i} \sim \Gamma^{-1}(3,3)$ where $l$ indexes layers, $i$ rows of the weight matrix, and $j$ columns.  This prior has two forms of Bayesian regularization.  The row-wise scale $\rlambda_{l, i}$ implements \textit{automatic relevance determination} (ARD) \citep{mackay1996bayesian, neal1994bayesian}, which controls the effective width.  The layer-wise scale $\rtau_{l}$ performs \textit{automatic depth determination} (ADD) \citep{nalisnick19dropout}, as it controls the effective depth.  We again compare three strategies for setting $\rtau$.  The first is to use a fixed scale ($\rtau = \tau_{0}$), thereby removing ADD and serving as a weak baseline.  The second is to use a shrinkage prior.  \citet{nalisnick19dropout} use an inverse gamma prior, and we report their results as the strong baseline.  For our method we use the PredCP with a log-Cauchy$(0,1)$ KLD prior.  For posterior inference, we use Bayes-by-backprop \citep{blundell2015weight} for the weights and variational EM \citep{wu2018fixingVB, nalisnick19dropout} for the scales $\rlambda$ and $\rtau$.  The maximization step cannot be performed analytically for the PredCP, as it can for the inverse gamma, and so we perform iterative gradient-based optimization.  Again, we use relatively small data sets to ensure the prior remains influential: results on UCI benchmarks \citep{Dua:2019, hernandez2015probabilistic} are reported in Table \ref{tab:UCIresults}.  Using the PredCP results in the best test set root-mean-square error (RMSE) for three of the seven benchmarks (\texttt{boston}, \texttt{energy}, \texttt{yacht}) and in one tie (\texttt{kin8nm}).

\paragraph{Few-Shot Learning}  Our final experiment evaluates the PredCP for few-shot learning.  We follow \citet{chen2019modular}'s experimental framework, using the hierarchical model $\mathcal{D}_{t} \sim p(\mathcal{D}_{t} | \rvtheta_{t}), \ \rvtheta_{t} \sim \text{N}(\vphi, \rtau \mathbb{I})$ (described in Section \ref{sec:apps}) and their $\sigma$-MAML algorithm for optimization.  In essence, $\sigma$-MAML performs MAP estimation for $\rvtheta_{t}$, $\vphi$, and $\rtau$.  The classifier is the standard four-layer convolutional NN \citep{finn2017model}.  We experimentally compare four different priors, each applied layer-wise (again following \citet{chen2019modular}).  The first baseline is $\rvtheta_{t} \sim \mathbbm{1}$ (improper uniform), which corresponds to standard MAML \citep{finn2017model}.  The second baseline is \citet{chen2019modular}'s model, which uses the improper uniform prior for the meta-parameters: $\vphi_{l}, \tau_{l} \sim \mathbbm{1}$.  For a third baseline, we extend \citet{chen2019modular}'s model by placing a shrinkage prior on $\rtau_{l}$, considering half-Cauchy, log-Cauchy, exponential, and gamma-exponential mixture distributions.  Finally, our proposal is to place a PredCP on $\rtau_{l}$.  We use the same four shrinkage priors for $\pi(\rkappa)$.  We evaluate all priors on the few-shot CIFAR100 \citep{oreshkin2018task} and mini-ImageNet \citep{vinyals2016matching} classification benchmarks, using the standard $5$-way $1$-shot and $5$-shot protocols.  Table \ref{tab:FSresults} reports the results.  The PredCP consistently improves accuracy across all experiments.

\section{Conclusions}
We proposed a novel prior termed the \textit{predictive complexity prior} (PredCP).  This prior is constructed procedurally and provides functional regularization.  We found the PredCP to improve generalization across a range of small-data tasks that require careful regularization.  The log-Cauchy$(0,1)$ served as a good default KLD prior.  One potential avenue for future work is to re-introduce predictive correlations into the divergence function, as these were lost when switching to the upper bound (Equation \ref{eq:KLD_upper}).  The resulting resnet-PredCP could then account for more interesting local structure in the predictive function, not just its point-wise variance.  Another direction is to explore other divergences, including symmetric ones such as the Hellinger distance.  Lastly, it would be interesting to investigate if the PredCP mitigates the misspecification issues described by \citet{wenzel2020good} or improves uncertainty estimation on out-of-distribution data.  


\bibliography{references}

\begin{thebibliography}{60}
\providecommand{\natexlab}[1]{#1}
\providecommand{\url}[1]{\texttt{#1}}
\expandafter\ifx\csname urlstyle\endcsname\relax
  \providecommand{\doi}[1]{doi: #1}\else
  \providecommand{\doi}{doi: \begingroup \urlstyle{rm}\Url}\fi

\bibitem[Alon et~al.(1999)Alon, Barkai, Notterman, Gish, Ybarra, Mack, and
  Levine]{alon1999broad}
Uri Alon, Naama Barkai, Daniel~A. Notterman, Kurt Gish, Suzanne Ybarra, Daniel
  Mack, and Arnold~J. Levine.
\newblock {Broad Patterns of Gene Expression Revealed by Clustering Analysis of
  Tumor and Normal Colon Tissues Probed by Oligonucleotide Arrays}.
\newblock In \emph{Proceedings of the National Academy of Sciences}, 1999.

\bibitem[Antorán et~al.(2020)Antorán, Allingham, and
  Hernández-Lobato]{antoran2020variational}
Javier Antorán, James~Urquhart Allingham, and José~Miguel Hernández-Lobato.
\newblock {Variational Depth Search in ResNets}.
\newblock \emph{ICLR Workshop on Neural Architecture Search}, 2020.

\bibitem[Atanov et~al.(2019)Atanov, Ashukha, Struminsky, Vetrov, and
  Welling]{atanov2018the}
Andrei Atanov, Arsenii Ashukha, Kirill Struminsky, Dmitriy Vetrov, and Max
  Welling.
\newblock {The Deep Weight Prior}.
\newblock In \emph{Proceedings of the International Conference on Learning
  Representations}, 2019.

\bibitem[Baldi and Sadowski(2013)]{baldi2013understanding}
Pierre Baldi and Peter~J. Sadowski.
\newblock {Understanding Dropout}.
\newblock In \emph{Advances in Neural Information Processing Systems}, 2013.

\bibitem[Bernardo(1979)]{bernardo1979reference}
Jose~M Bernardo.
\newblock {Reference Posterior Distributions for {B}ayesian Inference}.
\newblock \emph{Journal of the Royal Statistical Society. Series B
  (Methodological)}, 1979.

\bibitem[Bishop(2006)]{bishop2006pattern}
Christopher~M. Bishop.
\newblock \emph{{Pattern Recognition and Machine Learning}}.
\newblock Springer, 2006.

\bibitem[Blundell et~al.(2015)Blundell, Cornebise, Kavukcuoglu, and
  Wierstra]{blundell2015weight}
Charles Blundell, Julien Cornebise, Koray Kavukcuoglu, and Daan Wierstra.
\newblock {Weight Uncertainty in Neural Networks}.
\newblock In \emph{Proceedings of the 32nd International Conference on Machine
  Learning}, 2015.

\bibitem[Box(1980)]{box1980sampling}
George E.~P. Box.
\newblock {Sampling and Bayes' Inference in Scientific Modelling and
  Robustness}.
\newblock \emph{Journal of the Royal Statistical Society: Series A}, 1980.

\bibitem[Carpenter et~al.(2017)Carpenter, Gelman, Hoffman, Lee, Goodrich,
  Betancourt, Brubaker, Guo, Li, and Riddell]{carpenter2017stan}
Bob Carpenter, Andrew Gelman, Matthew~D. Hoffman, Daniel Lee, Ben Goodrich,
  Michael Betancourt, Marcus Brubaker, Jiqiang Guo, Peter Li, and Allen
  Riddell.
\newblock {Stan: A Probabilistic Programming Language}.
\newblock \emph{Journal of Statistical Software}, 2017.

\bibitem[Carvalho et~al.(2009)Carvalho, Polson, and
  Scott]{carvalho2009handling}
Carlos~M. Carvalho, Nicholas~G. Polson, and James~G. Scott.
\newblock {Handling Sparsity via the Horseshoe}.
\newblock In \emph{Proceedings of the 12th International Conference on
  Artificial Intelligence and Statistics}, 2009.

\bibitem[Casella(1985)]{casella1985introduction}
George Casella.
\newblock {An Introduction to Empirical Bayes Data Analysis}.
\newblock \emph{The American Statistician}, 1985.

\bibitem[Chen et~al.(2019)Chen, Friesen, Behbahani, Budden, Hoffman, Doucet,
  and de~Freitas]{chen2019modular}
Yutian Chen, Abram~L. Friesen, Feryal Behbahani, David Budden, Matthew~W.
  Hoffman, Arnaud Doucet, and Nando de~Freitas.
\newblock {Modular Meta-Learning with Shrinkage}.
\newblock \emph{NeurIPS Workshop on Meta-Learning}, 2019.

\bibitem[Dikov and Bayer(2019)]{dikov2019bayesian}
Georgi Dikov and Justin Bayer.
\newblock {Bayesian Learning of Neural Network Architectures}.
\newblock In \emph{Proceedings of the 22nd International Conference on
  Artificial Intelligence and Statistics}, 2019.

\bibitem[Dua and Graff(2019)]{Dua:2019}
Dheeru Dua and Casey Graff.
\newblock {UCI Machine Learning Repository}, 2019.

\bibitem[Finn et~al.(2017)Finn, Abbeel, and Levine]{finn2017model}
Chelsea Finn, Pieter Abbeel, and Sergey Levine.
\newblock {Model-Agnostic Meta-Learning for Fast Adaptation of Deep Networks}.
\newblock In \emph{Proceedings of the 34th International Conference on Machine
  Learning}, 2017.

\bibitem[Flam-Shepherd et~al.(2017)Flam-Shepherd, Requeima, and
  Duvenaud]{flam2017mapping}
Daniel Flam-Shepherd, James Requeima, and David Duvenaud.
\newblock {Mapping Gaussian Process Priors to Bayesian Neural Networks}.
\newblock \emph{NeurIPS Workshop on Bayesian Deep Learning}, 2017.

\bibitem[Flam-Shepherd et~al.(2018)Flam-Shepherd, Requeima, and
  Duvenaud]{flam2018characterizing}
Daniel Flam-Shepherd, James Requeima, and David Duvenaud.
\newblock {Characterizing and Warping the Function Space of Bayesian Neural
  Networks}.
\newblock \emph{NeurIPS Workshop on Bayesian Deep Learning}, 2018.

\bibitem[Fuglstad et~al.(2019)Fuglstad, Simpson, Lindgren, and
  Rue]{fuglstad2019constructing}
Geir-Arne Fuglstad, Daniel Simpson, Finn Lindgren, and H{\aa}vard Rue.
\newblock {Constructing Priors that Penalize the Complexity of Gaussian Random
  Fields}.
\newblock \emph{Journal of the American Statistical Association}, 2019.

\bibitem[Gal and Ghahramani(2016)]{gal2016dropout}
Yarin Gal and Zoubin Ghahramani.
\newblock {Dropout as a Bayesian Approximation: Representing Model Uncertainty
  in Deep Learning}.
\newblock In \emph{Proceedings of the 33rd International Conference on Machine
  Learning}, 2016.

\bibitem[Gelman(2006)]{gelman2006prior}
Andrew Gelman.
\newblock {Prior Distributions for Variance Parameters in Hierarchical Models}.
\newblock \emph{Bayesian Analysis}, 2006.

\bibitem[Gelman et~al.(2017)Gelman, Simpson, and Betancourt]{gelman2017prior}
Andrew Gelman, Daniel Simpson, and Michael Betancourt.
\newblock {The Prior Can Often Only Be Understood in the Context of the
  Likelihood}.
\newblock \emph{Entropy}, 2017.

\bibitem[Golub et~al.(1999)Golub, Slonim, Tamayo, Huard, Gaasenbeek, Mesirov,
  Coller, Loh, Downing, and Caligiuri]{golub1999molecular}
Todd~R. Golub, Donna~K. Slonim, Pablo Tamayo, Christine Huard, Michelle
  Gaasenbeek, Jill~P. Mesirov, Hilary Coller, Mignon~L. Loh, James~R. Downing,
  and Mark~A. Caligiuri.
\newblock {Molecular Classification of Cancer: Class Discovery and Class
  Prediction by Gene Expression Monitoring}.
\newblock \emph{Science}, 1999.

\bibitem[Hafner et~al.(2019)Hafner, Tran, Lillicrap, Irpan, and
  Davidson]{hafner2018noise}
Danijar Hafner, Dustin Tran, Timothy Lillicrap, Alex Irpan, and James Davidson.
\newblock {Noise Contrastive Priors for Functional Uncertainty}.
\newblock In \emph{Proceedings of the 35th Conference on Uncertainty in
  Artificial Intelligence}, 2019.

\bibitem[He et~al.(2016)He, Zhang, Ren, and Sun]{he2016Deep}
Kaiming He, Xiangyu Zhang, Shaoqing Ren, and Jian Sun.
\newblock {Deep Residual Learning for Image Recognition}.
\newblock In \emph{Proceedings of the IEEE {C}onference on {C}omputer {V}ision
  and {P}attern {R}ecognition}, 2016.

\bibitem[Heek and Kalchbrenner(2019)]{heek2019bayesian}
Jonathan Heek and Nal Kalchbrenner.
\newblock {Bayesian Inference for Large Scale Image Classification}.
\newblock \emph{ArXiv e-Prints}, 2019.

\bibitem[Hern{\'a}ndez-Lobato and Adams(2015)]{hernandez2015probabilistic}
Jos{\'e}~Miguel Hern{\'a}ndez-Lobato and Ryan Adams.
\newblock {Probabilistic Backpropagation for Scalable Learning of Bayesian
  Neural Networks}.
\newblock In \emph{Proceedings of the 32nd International Conference on Machine
  Learning}, 2015.

\bibitem[Jeffreys(1946)]{jeffreys1946invariant}
Harold Jeffreys.
\newblock {An Invariant Form for the Prior Probability in Estimation Problems}.
\newblock In \emph{Proceedings of the Royal Society of London A: Mathematical,
  Physical and Engineering Sciences}, 1946.

\bibitem[Johnson and Rossell(2010)]{johnson2010use}
Valen~E. Johnson and David Rossell.
\newblock {On the Use of Non-Local Prior Densities in Bayesian Hypothesis
  Tests}.
\newblock \emph{Journal of the Royal Statistical Society: Series B}, 2010.

\bibitem[Kingma and Ba(2014)]{kingma2014adam}
Diederik Kingma and Jimmy Ba.
\newblock {Adam: A Method for Stochastic Optimization}.
\newblock In \emph{Proceedings of the International Conference on Learning
  Representations}, 2014.

\bibitem[Kingma and Welling(2014)]{kingma2014efficient}
Diederik Kingma and Max Welling.
\newblock {Efficient Gradient-Based Inference Through Transformations Between
  Bayes Nets and Neural Nets}.
\newblock In \emph{Proceedings of the 31st International Conference on Machine
  Learning}, 2014.

\bibitem[Klein and Kneib(2016)]{klein2016scale}
Nadja Klein and Thomas Kneib.
\newblock {Scale-Dependent Priors for Variance Parameters in Structured
  Additive Distributional Regression}.
\newblock \emph{Bayesian Analysis}, 2016.

\bibitem[Kucukelbir et~al.(2017)Kucukelbir, Tran, Ranganath, Gelman, and
  Blei]{kucukelbir2017automatic}
Alp Kucukelbir, Dustin Tran, Rajesh Ranganath, Andrew Gelman, and David Blei.
\newblock {Automatic Differentiation Variational Inference}.
\newblock \emph{The Journal of Machine Learning Research}, 2017.

\bibitem[Louizos et~al.(2019)Louizos, Shi, Schutte, and
  Welling]{louizos2019functional}
Christos Louizos, Xiahan Shi, Klamer Schutte, and Max Welling.
\newblock {The Functional Neural Process}.
\newblock In \emph{Advances in Neural Information Processing Systems}, 2019.

\bibitem[Ma et~al.(2019)Ma, Li, and Hern{\'a}ndez-Lobato]{ma19implicit}
Chao Ma, Yingzhen Li, and Jos{\'e}~Miguel Hern{\'a}ndez-Lobato.
\newblock {Variational Implicit Processes}.
\newblock In \emph{Proceedings of the 36th International Conference on Machine
  Learning}, 2019.

\bibitem[Ma et~al.(2018)Ma, Ayaz, and Karaman]{ma2018invertibility}
Fangchang Ma, Ulas Ayaz, and Sertac Karaman.
\newblock {Invertibility of Convolutional Generative Networks from Partial
  Measurements}.
\newblock In \emph{Advances in Neural Information Processing Systems}, 2018.

\bibitem[MacKay(1994)]{mackay1996bayesian}
David MacKay.
\newblock {Bayesian Non-Linear Modeling for the Prediction Competition}.
\newblock \emph{Maximum Entropy and Bayesian Methods}, 1994.

\bibitem[Nalisnick and Smyth(2018)]{nalisnick2018learning}
Eric Nalisnick and Padhraic Smyth.
\newblock {Learning Priors for Invariance}.
\newblock In \emph{Proceedings of the 21st International Conference on
  Artificial Intelligence and Statistics}, 2018.

\bibitem[Nalisnick et~al.(2019)Nalisnick, Hern{\'a}ndez-Lobato, and
  Smyth]{nalisnick19dropout}
Eric Nalisnick, Jos{\'e}~Miguel Hern{\'a}ndez-Lobato, and Padhraic Smyth.
\newblock {Dropout as a Structured Shrinkage Prior}.
\newblock In \emph{Proceedings of the 36th International Conference on Machine
  Learning}, 2019.

\bibitem[Neal(1994)]{neal1994bayesian}
Radford~M. Neal.
\newblock \emph{{Bayesian Learning for Neural Networks}}.
\newblock PhD thesis, University of Toronto, 1994.

\bibitem[Oreshkin et~al.(2018)Oreshkin, Rodr\'{\i}guez~L\'{o}pez, and
  Lacoste]{oreshkin2018task}
Boris Oreshkin, Pau Rodr\'{\i}guez~L\'{o}pez, and Alexandre Lacoste.
\newblock {TADAM: Task Dependent Adaptive Metric for Improved Few-Shot
  Learning}.
\newblock In \emph{Advances in Neural Information Processing Systems}, 2018.

\bibitem[Papamakarios et~al.(2019)Papamakarios, Nalisnick, Rezende, Mohamed,
  and Lakshminarayanan]{papamakarios2019normalizing}
George Papamakarios, Eric Nalisnick, Danilo~Jimenez Rezende, Shakir Mohamed,
  and Balaji Lakshminarayanan.
\newblock {Normalizing Flows for Probabilistic Modeling and Inference}.
\newblock \emph{ArXiv e-Prints}, 2019.

\bibitem[Patr{\'\i}cio et~al.(2018)Patr{\'\i}cio, Pereira, Cris{\'o}stomo,
  Matafome, Gomes, Sei{\c{c}}a, and Caramelo]{patricio2018using}
Miguel Patr{\'\i}cio, Jos{\'e} Pereira, Joana Cris{\'o}stomo, Paulo Matafome,
  Manuel Gomes, Raquel Sei{\c{c}}a, and Francisco Caramelo.
\newblock {Using Resistin, Glucose, Age and BMI to Predict the Presence of
  Breast Cancer}.
\newblock \emph{BMC Cancer}, 2018.

\bibitem[Pearce et~al.(2019)Pearce, Tsuchida, Zaki, Brintrup, and
  Neely]{pearce2019expressive}
Tim Pearce, Russell Tsuchida, Mohamed Zaki, Alexandra Brintrup, and Andy Neely.
\newblock {Expressive Priors in Bayesian Neural Networks: Kernel Combinations
  and Periodic Functions}.
\newblock In \emph{Proceedings of the 35th Conference on Uncertainty in
  Artificial Intelligence}, 2019.

\bibitem[Phuong and Lampert(2020)]{phuong2020functional}
Mary Phuong and Christoph~H. Lampert.
\newblock {Functional vs. Parametric Equivalence of ReLU Networks}.
\newblock In \emph{Proceedings of the International Conference on Learning
  Representations}, 2020.

\bibitem[Piironen and Vehtari(2017{\natexlab{a}})]{piironen2017hyperprior}
Juho Piironen and Aki Vehtari.
\newblock {On the Hyperprior Choice for the Global Shrinkage Parameter in the
  Horseshoe Prior}.
\newblock In \emph{Proceedings of the 20th International Conference on
  Artificial Intelligence and Statistics}, 2017{\natexlab{a}}.

\bibitem[Piironen and Vehtari(2017{\natexlab{b}})]{piironen2017sparsity}
Juho Piironen and Aki Vehtari.
\newblock {Sparsity Information and Regularization in the Horseshoe and Other
  Shrinkage Priors}.
\newblock \emph{Electronic Journal of Statistics}, 2017{\natexlab{b}}.

\bibitem[Robert(2001)]{robert2007bayesian}
Christian Robert.
\newblock \emph{{The Bayesian Choice}}.
\newblock Springer, 2001.

\bibitem[Shin et~al.(2018)Shin, Bhattacharya, and Johnson]{shin2018scalable}
Minsuk Shin, Anirban Bhattacharya, and Valen~E. Johnson.
\newblock {Scalable Bayesian Variable Selection Using Nonlocal Prior Densities
  in Ultrahigh-Dimensional Settings}.
\newblock \emph{Statistica Sinica}, 2018.

\bibitem[Simpson et~al.(2017)Simpson, Rue, Riebler, Martins, and
  S{\o}rbye]{simpson2017penalising}
Daniel Simpson, H{\aa}vard Rue, Andrea Riebler, Thiago~G. Martins, and
  Sigrunn~H. S{\o}rbye.
\newblock {Penalising Model Component Complexity: A Principled, Practical
  Approach to Constructing Priors}.
\newblock \emph{Statistical Science}, 2017.

\bibitem[Song et~al.(2019)Song, Meng, and Ermon]{song2019mintnet}
Yang Song, Chenlin Meng, and Stefano Ermon.
\newblock {MintNet: Building Invertible Neural Networks with Masked
  Convolutions}.
\newblock In \emph{Advances in Neural Information Processing Systems}, 2019.

\bibitem[S{\o}rbye and Rue(2017)]{sorbye2017penalised}
Sigrunn~Holbek S{\o}rbye and H{\aa}vard Rue.
\newblock {Penalised Complexity Priors for Stationary Autoregressive
  Processes}.
\newblock \emph{Journal of Time Series Analysis}, 2017.

\bibitem[Sun et~al.(2019)Sun, Zhang, Shi, and Grosse]{sun2018functional}
Shengyang Sun, Guodong Zhang, Jiaxin Shi, and Roger Grosse.
\newblock {Functional Variational Bayesian Neural Networks}.
\newblock In \emph{Proceedings of the International Conference on Learning
  Representations}, 2019.

\bibitem[Ventrucci and Rue(2016)]{ventrucci2016penalized}
Massimo Ventrucci and H{\aa}vard Rue.
\newblock {Penalized Complexity Priors for Degrees of Freedom in Bayesian
  P-Splines}.
\newblock \emph{Statistical Modelling}, 2016.

\bibitem[Ventrucci et~al.(2019)Ventrucci, Cocchi, Burgazzi, and
  Laini]{ventrucci2019pc}
Massimo Ventrucci, Daniela Cocchi, Gemma Burgazzi, and Alex Laini.
\newblock {PC Priors for Residual Correlation Parameters in One-Factor Mixed
  Models}.
\newblock \emph{Statistical Methods \& Applications}, 2019.

\bibitem[Vinyals et~al.(2016)Vinyals, Blundell, Lillicrap, and
  Wierstra]{vinyals2016matching}
Oriol Vinyals, Charles Blundell, Tim Lillicrap, and Daan Wierstra.
\newblock {Matching Networks for One Shot Learning}.
\newblock In \emph{Advances in Neural Information Processing Systems}, 2016.

\bibitem[Wen et~al.(2018)Wen, Vicol, Ba, Tran, and Grosse]{wen2018flipout}
Yeming Wen, Paul Vicol, Jimmy Ba, Dustin Tran, and Roger Grosse.
\newblock {Flipout: Efficient Pseudo-Independent Weight Perturbations on
  Mini-Batches}.
\newblock In \emph{Proceedings of the International Conference on Learning
  Representations}, 2018.

\bibitem[Wenzel et~al.(2020)Wenzel, Roth, Veeling, {\'S}wi{\k{a}}tkowski, Tran,
  Mandt, Snoek, Salimans, Jenatton, and Nowozin]{wenzel2020good}
Florian Wenzel, Kevin Roth, Bastiaan~S. Veeling, Jakub {\'S}wi{\k{a}}tkowski,
  Linh Tran, Stephan Mandt, Jasper Snoek, Tim Salimans, Rodolphe Jenatton, and
  Sebastian Nowozin.
\newblock {How Good is the Bayes Posterior in Deep Neural Networks Really?}
\newblock In \emph{Proceedings of the 37th International Conference on Machine
  Learning}, 2020.

\bibitem[{Wu} et~al.(2019){Wu}, {Nowozin}, {Meeds}, {Turner},
  Hern{\'a}ndez-Lobato, and {Gaunt}]{wu2018fixingVB}
Anqi {Wu}, Sebastian {Nowozin}, Edward {Meeds}, Richard~E. {Turner},
  Jos{\'e}~Miguel Hern{\'a}ndez-Lobato, and Alexander~L. {Gaunt}.
\newblock {Deterministic Variational Inference for Robust Bayesian Neural
  Networks}.
\newblock In \emph{Proceedings of the International Conference on Learning
  Representations}, 2019.

\bibitem[Zellner(1986)]{zellner1986gPrior}
A.~Zellner.
\newblock {On Assessing Prior Distributions and Bayesian Regression Analysis
  with g-Prior Distributions}.
\newblock \emph{Bayesian Inference and Decision Techniques}, 1986.

\bibitem[Zhang et~al.(2020)Zhang, Li, Zhang, Chen, and
  Wilson]{Zhang2020Cyclical}
Ruqi Zhang, Chunyuan Li, Jianyi Zhang, Changyou Chen, and Andrew~Gordon Wilson.
\newblock {Cyclical Stochastic Gradient MCMC for Bayesian Deep Learning}.
\newblock In \emph{Proceedings of the International Conference on Learning
  Representations}, 2020.

\end{thebibliography}
\bibliographystyle{plainnat}

\appendix
\onecolumn
\section{ECP FOR LINEAR REGRESSION, CONTINUED}\label{app:lr_cont}

\subsection{Varying Prior Parameters}
We consider the following priors on the KLD, denoted $\pi(\rkappa)$ in the main text:
\begin{itemize}
    \item $\text{Exponential}(x; \lambda) = e^{-x/\lambda} / \lambda$
    \item $\text{Gamma}(x; \vlambda = \{\lambda_{1}, \lambda_{2}\}) = \frac{x^{\lambda_{1} - 1}}{\Gamma(\lambda_{1})\lambda_{2}} e^{-x/\lambda_{2}} $
    \item $\text{Log-Cauchy}(x;\lambda) =  \frac{1}{\pi x} \left(\frac{\lambda}{(\log x)^{2} + \lambda^{2}} \right)$
\end{itemize}  Since the gamma prior favors $p_{0}$ and the exponential $p_{+}$, we also consider a mixture of the two: \begin{equation*}
    \pi( \rkappa; \lambda) = \lambda \text{Gamma}\left( \text{KL}\left[ p_{+} \ || \  p_{0} \right] \right) + (1-\lambda)\text{Exponential}\left( \text{KL}\left[ p_{+} \ || \  p_{0} \right] \right).
\end{equation*}  This mixture should achieve the same interpolation behavior as the log-Cauchy, balancing preferences for $p_{0}$ and $p_{+}$.  See Figure \ref{subfig:lr_tau_mix} for a visualization of the ECP for the gamma-exponential mixture as the mixing coefficient is varied.  Figure \ref{subfig:lr_tau_varLam} shows the log-Cauchy ECP as its scale is varied.

\subsection{Dynamic Shrinkage Profile} 
\begin{figure}[h]
\centering
\subfigure[Varying $\lambda$ for Gam-Exp Mixture]{
\includegraphics[width=0.3\linewidth]{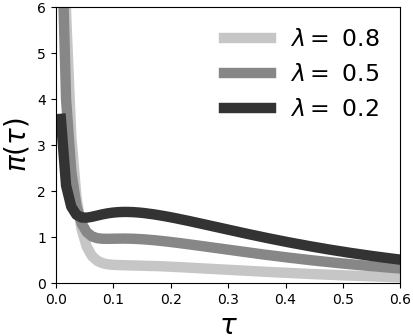}\label{subfig:lr_tau_mix}}
\hfill
\subfigure[Varying $\lambda$ for Log-Cauchy]{
\includegraphics[width=0.3\linewidth]{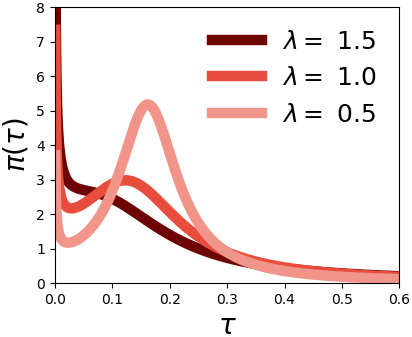}\label{subfig:lr_tau_varLam}}
\hfill 
\subfigure[Log-Cauchy Shrinkage Profile]{
\includegraphics[width=0.3\linewidth]{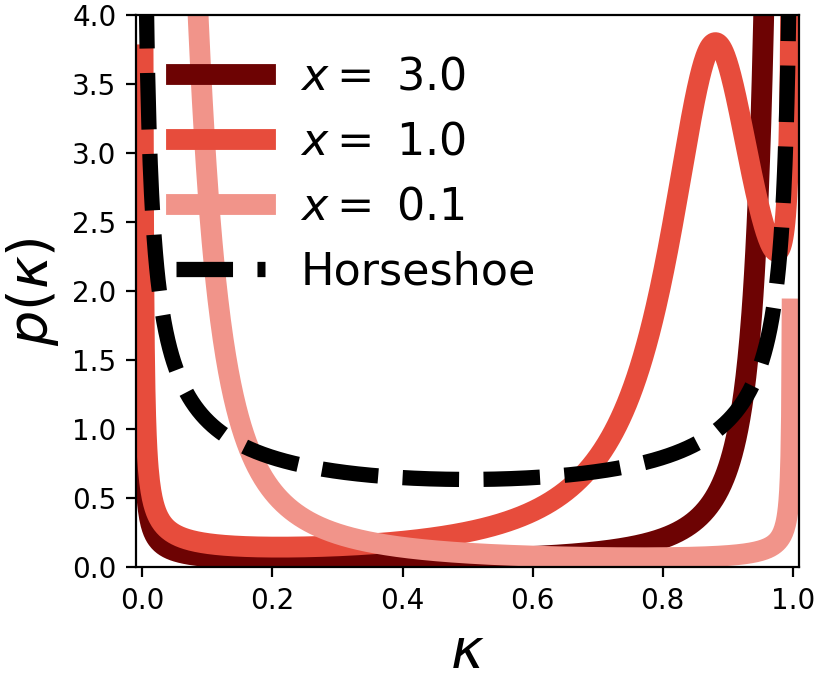}\label{subfig:lc_shrinks}}
\subfigure[Mixture Shrinkage Profile]{
\includegraphics[width=0.3\linewidth]{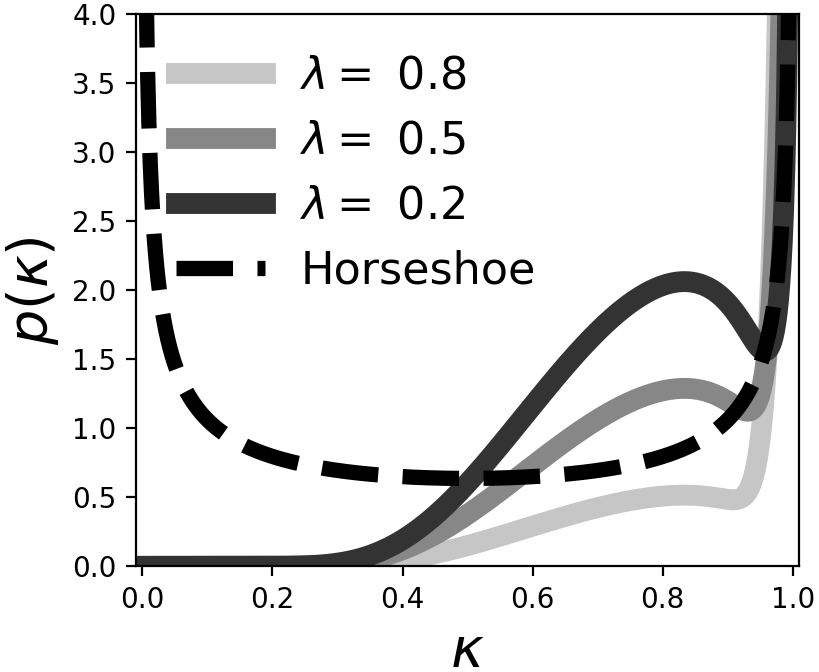}\label{subfig:mix_shrinks}}
\hfill
\subfigure[Varying Rank of $\mX$]{
\includegraphics[width=0.3\linewidth]{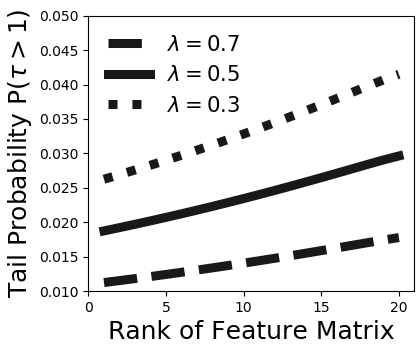}\label{subfig:tail_vs_rank}}
\hfill 
\subfigure[Varying Scale for Full-Rank $\mX$]{
\includegraphics[width=0.321\linewidth]{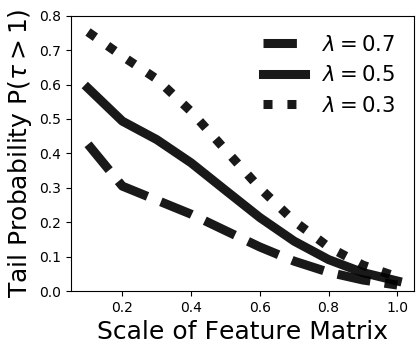}\label{subfig:tail_vs_scale_full}}
\caption{\textit{$p(\rtau)$ for Linear Regression}.  Subfigures \ref{subfig:lr_tau_mix} and \ref{subfig:lr_tau_varLam} show how the mixture and log-Cauchy ECP for $\rtau$ changes as a function of $\lambda$.  Subfigures \ref{subfig:lc_shrinks} and \ref{subfig:mix_shrinks} show the shrinkage profiles for the log-Cauchy and mixture ECPs (respectively).  Subfigures \ref{subfig:tail_vs_rank} and \ref{subfig:tail_vs_scale_full} show how varying the rank and scale of the design matrix affects the ECP's tail probability.  
}
\label{fig:ppcp_lrShrink2}
\end{figure}

One method for illuminating and comparing the effects of shrinkage priors is through their corresponding \textit{shrinkage profile} \citep{carvalho2009handling}.  Consider the model: $\ry \sim \text{N}(\rtheta, 1)$, $\rtheta \sim \text{N}(0, \tau)$, $\tau \sim p(\tau)$.  Given one observation $\ry = y_{0}$, the Bayes estimator for $\rtheta$ is: $\mathbb{E}[\rtheta | y_{0}, \tau] = \kappa \cdot 0 + (1-\kappa) \cdot y_{0}$, $\kappa = 1 /(1 + \tau)$.  Making the change of variables $p(\kappa) = p(\tau) | \frac{d}{d \kappa} 1/\kappa - 1 |$, we can examine the induced prior $p(\kappa)$.  If $p(\kappa)$ places high density near $\kappa=1$, then the prior provides strong regularization since the Bayes estimator would be near zero.  Conversely, a strong mode at $\kappa=0$ means the resulting estimator would be near $y_{0}$, implying a tendency to follow the data.  Figure \ref{subfig:lc_shrinks} shows the shrinkage profile for the log-Cauchy ECP for $x\in\{.1, 1, 3 \}$, comparing the profiles of the horseshoe's.  The horseshoe is an effective prior for robust regression since it is designed to equally favor $0$ and $1$ ($\rkappa \sim $ Beta$(.5, .5)$).  The shrinkage profile for the log-Cauchy also can place density at both extremes.  However, unlike the horseshoe, the ECP enables \emph{dynamic} shrinkage, being able to adjust the profile as a function of $x$.  We see that for $x=0.1$, the log-Cauchy ECP actually favors the unshrunk solution whereas for $x=3$ strong shrinkage is preferred.  The mixture ECP's shrinkage profile is shown in Figure \ref{subfig:mix_shrinks}.  Due to the exponential not having a heavy tail, the mixture ECP's profile cannot place any significant density at $\kappa = 0$, meaning that the model can never completely `forget' the shrinkage regularization.

\subsection{Example: Multivariate Regression}  
We now examine the case of multivariate regression: $ \mathbb{E}[\rvy | \mX, \rvbeta] = \mX \rvbeta$, where $\rvy \in \mathbb{R}^{N}$ is an $N$-dimensional vector of responses, $\mX \in \mathbb{R}^{N \times D}$ the design matrix, and $\rvbeta \in \mathbb{R}^{D}$ a vector of parameters.  Using the multivariate analogs of the priors from above---$p_{0} = \delta(|\rvbeta - \mathbf{0} |)$, \ $p_{+} = \text{N}(\mathbf{0}, \rtau \mathbb{I})$---we can derive the following ECP: \begin{equation}
    p(\rtau; \vlambda, \mX) = \pi_{\text{\tiny{KL}}}\left( \frac{-1}{2} \log | \mathbb{I} + \rtau \sigma^{-2}_{y} \mX^{T} \mX |  + \frac{\rtau}{2\sigma^{2}_{y}}\text{tr}\left\{ \mX^{T} \mX \right\} ; \vlambda \right) \  \left| \frac{\partial  \ \text{KL}}{\partial \ \rtau} \right| \end{equation} 
where $\text{tr}\{ \cdot \}$ denotes the trace operation and $| \cdot |$ the determinant.  The KLD is computed between $p(\rvy | \mX, \rtau)$ and $p_{0}(\rvy | \mX)$.  We omit the details of the volume term due to space constraints.  

    
Here the KLD is a multidimensional integral that takes into account \emph{correlations} in the model's predictions.  Hence, we can explore how characteristics of the design matrix influence the prior.  We consider the tail probability $P(\rtau > 1)$ since, as probability mass moves away from the origin, the ECP increasingly prefers the extended model.  Below we describe simulations using the mixture ECP due to its mixing weight $\lambda$ being interpretable.  First consider the case in which $\mX$ has a rank of one and all row vectors have a length of one.  This means that the data is essentially redundant, generating the same predictions.  As the rank of $\mX$ increases, the predictions start to become varied and the model output becomes responsive to each $\vx_{n}$.  Thus, we should expect $p(\rtau)$ to favor larger values as the rank of $\mX$ increases.  Indeed, this is exactly the behavior we observe in Figure \ref{subfig:tail_vs_rank}, plotting the tail probability $P(\rtau > 1)$ as we vary the rank from $1$ to $20$.  Another attribute of $\mX$ we can vary is its scale: $\alpha \mX$.  Subfigure \ref{subfig:tail_vs_scale_full} shows the tail probability as the scale $\alpha$ increases (for full-rank $\mX$).  We find that scale changes result in more pronounced tail effects than changes in rank.

\subsection{Example: ECP for the Linear Regression Coefficient}  
\begin{figure}
\centering
\subfigure[Varying $\pi_{\text{\tiny{KL}}}$]{
\includegraphics[width=0.3\linewidth]{KLD_priors.png}\label{subfig:lr_KLD}}
\hfill
\subfigure[Induced Prior $\pi(\beta_{1})$]{
\includegraphics[width=0.3\linewidth]{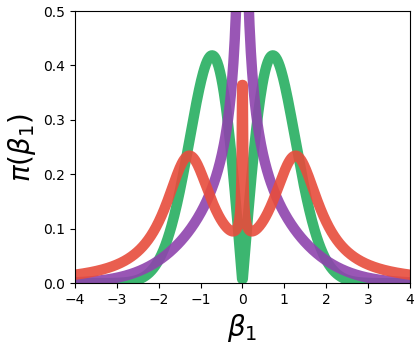}\label{subfig:lr_beta1}}
\hfill 
\subfigure[Varying $\rx$ for Log-Cauchy]{
\includegraphics[width=0.3\linewidth]{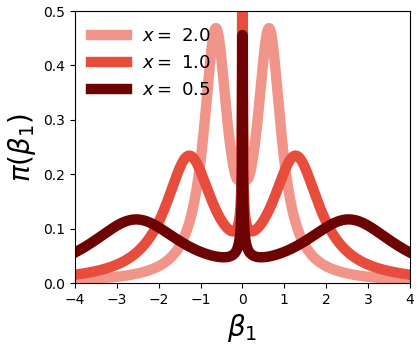}\label{subfig:lr_varX}}
\hfill
\subfigure[Varying $\lambda$ for Log-Cauchy]{
\includegraphics[width=0.3\linewidth]{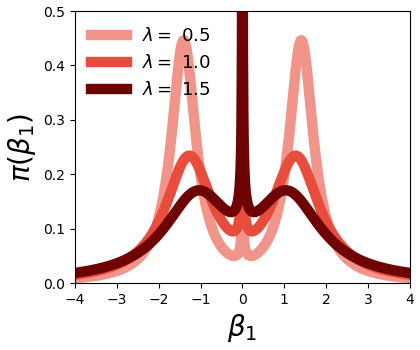}\label{subfig:lr_varLam}}
\hfill 
\subfigure[Varying $\lambda$ for Gam-Exp Mixture]{
\includegraphics[width=0.3\linewidth]{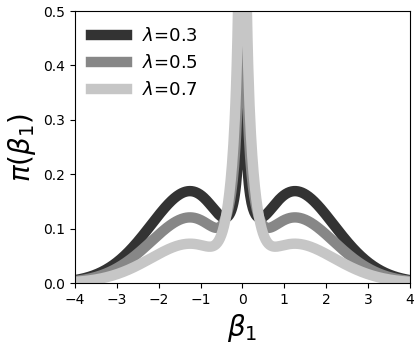}\label{subfig:lr_mix_varLam}}
\hfill 
\subfigure[Varying $\sigma$ for Non-Local Prior]{
\includegraphics[width=0.3\linewidth]{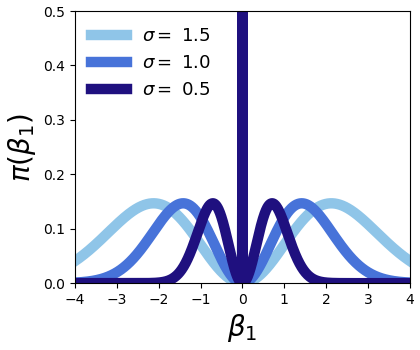}\label{subfig:lr_nonLocal}}
\caption{\textit{$\pi(\beta_{1})$ for Linear Regression}.  Subfigure \ref{subfig:lr_KLD} shows three choices for $\pi_{\text{\tiny{KL}}}$: exponential ($\lambda = .5$), gamma ($\lambda = (.2, 2)$), and log-Cauchy ($\lambda = 1$).  Subfigure \ref{subfig:lr_beta1} shows the priors induced on $\beta_{1}$ for each KLD prior.  Subfigure \ref{subfig:lr_varX} shows how the log-Cauchy ECP changes as a function of $x$.  Subfigure \ref{subfig:lr_varLam} shows how the log-Cauchy ECP changes when varying its scale parameter ($x = 1$). Subfigure \ref{subfig:lr_mix_varLam} shows how the mixture (gamma and exponential) ECP changes when varying the mixture weight $\lambda$ ($x = 1$).  Subfigure \ref{subfig:lr_nonLocal} shows the non-local prior for three scales.  
}
\label{fig:ppcp_lr}
\end{figure}

In the main text, we consider applying the ECP to the scale of the prior on $\rvbeta$, the regression coefficients.  Yet, we can also define the ECP on $\rvbeta$ directly. Consider the linear model $ \mathbb{E}[\ry | x, \rvbeta] = \beta_{0} + \beta_{1}x$, $\ \beta_{0} \sim \text{N}(0, \sigma_{\beta_{0}}^{2})$.  We wish a define the ECP on $\beta_{1}$ to control deviation from the base model $p_{0}: \beta_{1} = 0$: \begin{equation}\begin{split}
    p(\beta_{1}; \lambda, x) &= \pi_{\text{\tiny{KL}}}\left( \text{KL}\left[ p_{+}(\ry | x, \beta_{1}) \ || \  p_{0}(\ry | x) \right]; \lambda \right) \left| \frac{\partial  \ \text{KL}\left[ p_{+}(\ry | x, \beta_{1}) \ || \  p_{0}(\ry | x) \right]}{\partial \beta_{1}} \right| \\ &= \frac{1}{2} \  \pi_{\text{\tiny{KL}}}\left( \frac{\beta_{1}^{2}x^{2}}{2(\sigma^{2}_{y} + \sigma^{2}_{\beta_{0}})} ; \lambda \right) \left| \frac{\beta_{1}x^{2}}{\sigma^{2}_{y} + \sigma^{2}_{\beta_{0}}} \right|
\end{split}
\end{equation} where $\sigma^{2}_{y}$ is the response noise.  Figure \ref{subfig:lr_KLD} shows three choices for $\pi_{\text{\tiny{KL}}}$---exponential (green), gamma (purple), and log-Cauchy (red)---and Figure \ref{subfig:lr_beta1} shows the prior each induces on $\beta_{1}$.  We see that choice of $\pi_{\text{\tiny{KL}}}$ is significant.  If $\pi_{\text{\tiny{KL}}}$ places too little density at zero, the volume term $| \beta_{1}x^{2} / (\sigma^{2}_{y} + \sigma^{2}_{\beta_{0}})|$ in the ECP dominates, driving $p(\beta_{1})$ to zero at $\beta_{1} = 0$.  In turn, this drives $\beta_{1}$ from reflecting the behavior of $p_{0}$ and suppresses any regularization.  We see this behavior from the exponential (green) as its ECP has no density at zero.  At the other end of the spectrum is the gamma (purple).  It places too much density at zero, forcing $p(\beta_{1})$ to preference the base model $\beta_{1}=0$.  Lastly, the log-Cauchy (red) has the most interesting behavior: it has strong shrinkage at the origin to reflect $p_{0}$ but also significant density away from zero to represent $p_{+}$.  Hence the log-Cauchy is able to balance preferences for both $p_{+}$ and $p_{0}$, interpolating between the exponential and gamma's single-mindedness.   Lastly, Figure \ref{subfig:lr_varX} shows how $p(\beta_{1})$ changes as $x$ is varied---the data adaptive nature mentioned above.  The ECP's shrinkage is adjusted to the scale of the features, applying stronger regularization when $x$ is large and relaxing as $x$ decreases.  This is sensible since the models predictions cannot change as drastically for small $x$s as they can for large $x$s.


\paragraph{Connection to Non-Local Priors}  Perhaps the (log-Cauchy and mixture) ECP's most interesting feature is in how it balances between $p_{0}$ and $p_{+}$ through multi-modality.  In addition to the strong mode at zero, there are modes separated from and symmetric about the origin; see Figure \ref{subfig:lr_varLam}.  This is the defining feature of so-called \textit{non-local priors} \citep{johnson2010use}.  This class of priors is designed to achieve good convergence rates in Bayesian hypothesis testing by carefully allocating density exactly at the null and distinctly away from the null to represent the alternative.  \citet{shin2018scalable} apply this non-local principle to Bayesian variable selection in high-dimensional regression via the following prior: \begin{equation}
    \beta \sim \rz \delta_{0} + (1-\rz) p_{\text{\tiny{NL}}}(\beta), \ \ \ \ \rz \sim \text{Bernoulli}(\rho), \ \ \ \ p_{\text{\tiny{NL}}}(\beta) = \frac{\beta^{2}}{\sigma} \text{N}(\beta; 0, \sigma)
\end{equation} where $\beta$ is the linear model's coefficient and $p_{\text{\tiny{NL}}}$ is known as a \textit{product (2nd) moment prior} \citep{johnson2010use}.  See Figure \ref{subfig:lr_nonLocal} for a visualization.  While the ECP and non-local prior have clear similarities, the connection can be made explicit my considering the Bayes factor $\text{BF}(+ | 0) = p_{+}(\ry | \vx, \rtheta_{+}) / p_{0}(\ry | \vx)$, which is what we would use to test $\text{H}_{0}: \rtheta_{+}=0$ vs $\text{H}_{1}: \rtheta_{+} \ne 0$.  The ECP's KLD term contains this exact expression: $ \text{KL}\left[ p_{+} \ || \  p_{0} \right] =  \mathbb{E}_{p_{+}}\left[ \log \text{BF}(+ | 0) \right], $ which can be interpreted as the (log) Bayes factor expected if the data is truly generated by the extended model.  


\section{BIJECTIVITY CONDITIONS FOR NEURAL NETWORKS}\label{app:conds}
Below we show that $\mathbb{E}_{\rvtheta | \tau}\text{KL}[p_{+} || p_{0}]$ is bijective w.r.t.~$\tau$ for Gaussian and categorical observation models, ReLU activations, and Gaussian weight priors $\rmW_{l} \sim \text{N}(\mPhi_{l}, \rtau_{l} \mSigma)$.  We assume the base model has weights $\rmW_{l} \sim \delta[\mPhi_{l}]$, where $\mPhi_{l}$ is the prior mean of the expanded model.  In turn, the change of variables is well-defined and the PredCP is proper (i.e.~integrates to $1$) when used in the applications given in Section 5.  Before diving into the technical details, we point out that invertibility should not be hard to guarantee since $\rtau \in \mathbb{R}^{+}$ is a scalar.  If the reader is familiar with the literature on \textit{normalizing flows} \citep{papamakarios2019normalizing} and invertible architectures \citep{song2019mintnet}, one may have the impression that invertibility is hard to guarantee for neural networks, often requiring constraints on the weights.  This is only the case because an invertible architecture must preserve bijectivity w.r.t.~the entire input \emph{vector}.  We, to the contrary, just have to preserve scalar information.  Moreover, we know this scalar $\rtau$ is strictly positive, which eliminates any symmetry about the origin.  Thus, intuitively, the information about $\rtau$ should be preserved at every hidden layer as long as at least one ReLU unit is active.  This brings up the only assumption that we require: \textit{non-degeneracy}.  That is, for every hidden layer, there must be at least one active ReLU unit (i.e.~a unit evaluated to a positive value).  This is an extremely weak assumption, especially for all but the smallest neural networks, and ReLU networks commonly satisfy much stronger non-degeneracy assumptions \citep{phuong2020functional}.  If dead layers would arise for some reason, initializing the biases to be positive should prevent the pathology.  Before moving on to the main results, we first introduce two conventions.  

\paragraph{Non-Centered Parameterization:} We assume all weights are represented in the observation model $p(\rvy | \mX, \rmW)$ using the following non-centered form: $\rmW_{l} \overset{d}{=} \mPhi_{l} + \sqrt{\rtau_{l}} \cdot \rmXi$, $\rmXi \sim \text{N}(\mathbf{0}, \mSigma)$.  This parameterization is useful for `exposing' $\rtau_{l}$ so that we can write the KLD as an explicit function of $\rtau$.

\paragraph{Linear Representation of ReLU Activations}  Recall that feedforward networks with ReLU activations are piecewise-linear functions.  Thus, it is equivalent to express a dense ReLU layer in terms of a diagonal `gating' matrix $\tilde{\mathbb{I}}$ \citep{ma2018invertibility}: \begin{equation}
    \text{\texttt{ReLU}}(  \rvh \rmW ) = \rvh \rmW \tilde{\mathbb{I}}, \ \ \ \text{ where } \ \ \ \tilde{\ri}_{j, j} = 1 \ \  \text{ if } \ \ 0 < \sum_{i=1}^{D}  \rw_{i,j} \rh_{i},
\end{equation} otherwise $\tilde{\ri}_{j, j} = 0$.  Using this convention, we can then represent the network's linear output as: \begin{equation*}
    \rmF_{out} = \mX  \left(\prod_{l=1}^{L} \rmW_{l} \tilde{\mathbb{I}}_{l} \right) \mW_{out}
\end{equation*} where $\rmF_{out}$ is a matrix containing the network's linear output (i.e.~before a softmax is applied, in the classification setting) for all training features $\mX$.

\subsection{Gaussian Observation Model}
Let's now consider checking for bijectivity w.r.t.~$\tau_{l}$ when the observation model is Gaussian, i.e. $\rvy \sim \text{N}(\rvf_{out}, \sigma^{2}_{y}\mathbb{I})$.  The expected KLD is then: \begin{equation}\label{expKL_1}\begin{split}
    \mathbb{E}_{\rmW_{l} | \tau_{l}}&\text{KL}[\text{N}(\rmF_{out}^{+}, \sigma^{2}_{y}\mathbb{I}) ||  \text{N}(\mF_{out}^{0}, \sigma^{2}_{y}\mathbb{I})] = \frac{1}{2\sigma^{2}_{y}} \sum_{n=1}^{N}\sum_{d=1}^{D} \mathbb{E}_{\rmW_{l} | \tau_{l}}\left[\left(\rf_{n,d, out}^{+} - f_{n,d,out}^{0}\right)^{2}\right] \\ &= \frac{1}{2\sigma^{2}_{y}} \sum_{n=1}^{N}\sum_{d=1}^{D}  \mathbb{E}_{\rmW_{l} | \tau_{l}}[(\rf^{+}_{n,d,out})^{2}] - 2 \mathbb{E}_{\rmW_{l} | \tau_{l}}[\rf^{+}_{n,d,out}] f_{n,d,out}^{0} + (f_{n,d,out}^{0})^{2}
\end{split}\end{equation} where $n$ indexes the training features and $d$ the output dimensions.  Since $f_{out}^{0}$ does not involve $\tau_{l}$, we can treat is as a constant.  The sum over $n$ and $d$ forms a conic combination.  Therefore if all terms have the same monotonicity and are bijective, then the sum will be bijective as well. 

We first turn toward the expectation in the middle term and expand it using the non-centered parameterization.  We drop the indexes to help with notational clutter: \begin{equation}\label{expan1}\begin{split}
    \mathbb{E}_{\rmW_{l} | \tau_{l}}[f^{+}_{out}] &= \mathbb{E}_{\rmW_{l} | \tau_{l}}\left[ \vx \left(\prod_{l=1}^{L} \tilde{\mathbb{I}}_{l} \rmW_{l}\right) \vw_{out}  \right] \\ &= \mathbb{E}_{\rmXi_{l}}\left[ \vx \left(\prod_{l=1}^{L}  (\mPhi_{l} + \sqrt{\tau_{l}} \cdot \rmXi_{l}) \tilde{\mathbb{I}}_{l} \right) \vw_{out} \right]
    \\ &= \mathbb{E}_{\rmXi_{l}}\left[ \vx \left(\prod_{l=1}^{L}  \mPhi_{l} \tilde{\mathbb{I}}_{l} \right) \vw_{out} \right] + \mathbb{E}_{\rmXi_{l}}\left[ \vx \left(\prod_{l=1}^{L}  \sqrt{\tau_{l}} \cdot \rmXi_{l} \tilde{\mathbb{I}}_{l} \right) \vw_{out} \right].
\end{split}\end{equation}  Pushing through the expectation, we have: \begin{equation*}\begin{split}
&=  \vx \left(\prod_{l=1}^{L}  \mPhi_{l} \mathbb{E}_{\rmXi_{l}}\left[\tilde{\mathbb{I}}_{l}\right] \right) \vw_{out}  +  \vx \left(\prod_{l=1}^{L}  \sqrt{\tau_{l}} \cdot \mathbb{E}_{\rmXi_{l}}\left[\rmXi_{l} \tilde{\mathbb{I}}_{l} \right] \right) \vw_{out}.
\end{split}\end{equation*}  Considering the first term, we have that $\mathbb{E}[\tilde{i}] = 1$ if $\sum_{i} \phi_{i,j} \rh_{i} > 0$.  For the second, $\mathbb{E}_{\rmXi_{l}}\left[\rmXi_{l} \tilde{\mathbb{I}}_{l} \right] = \mathbf{0} $.  Thus, $\mathbb{E}_{\rmW_{l} | \tau_{l}}[f^{+}_{out}]$ reduces to the first term, and since this term depends only on the prior means, it is equivalent to the output of the base model:
\begin{equation}
     \mathbb{E}_{\rmW_{l} | \tau_{l}}[f^{+}_{out}]  = f^{0}_{out}.
\end{equation}

Now turning to the other expectation term in Equation \ref{expKL_1} and using the expansion from Equation \ref{expan1}, we have: \begin{equation*}\begin{split}
    \mathbb{E}_{\rmW_{l} | \tau_{l}} & [(f^{+}_{out})^{2}] = \\ & \mathbb{E}_{\rmXi_{l}}\Bigg[ \left( \vx \left(\prod_{l=1}^{L}  \mPhi_{l} \tilde{\mathbb{I}}_{l} \right) \vw_{out} \right)^{2} + 2\left(\vx \left(\prod_{l=1}^{L}  \mPhi_{l} \tilde{\mathbb{I}}_{l} \right) \vw_{out} \right) \left( \vx \left(\prod_{l=1}^{L}  \sqrt{\tau_{l}} \cdot \rmXi_{l} \tilde{\mathbb{I}}_{l} \right) \vw_{out} \right) \\ &+ \left(\vx \left(\prod_{l=1}^{L} \sqrt{\tau_{l}} \cdot \rmXi_{l} \tilde{\mathbb{I}}_{l}  \right) \vw_{out} \right)^{2}  \Bigg].
\end{split}\end{equation*}  The middle term drops out after taking the expectation.  We are left with the two remaining terms: \begin{equation*}
    \mathbb{E}_{\rmW_{l} | \tau_{l}}[(f^{+}_{out})^{2}] = \underbrace{\mathbb{E}_{\rmXi_{l}}\left[ \left( \vx \left(\prod_{l=1}^{L}  \mPhi_{l} \tilde{\mathbb{I}}_{l} \right) \vw_{out} \right)^{2} \right]}_{\gamma_{1}^{\scriptscriptstyle{> 0}}} + \tau_{l} \cdot \underbrace{\prod_{l' \ne l} \tau_{l'} \cdot \mathbb{E}_{\rmXi_{l}}\left[ \left(\vx \left(\prod_{l=1}^{L}  \rmXi_{l} \tilde{\mathbb{I}}_{l} \right) \vw_{out} \right)^{2} \right]}_{\gamma_{2}^{\scriptscriptstyle{> 0}} }. 
\end{equation*} We use the variables $\gamma_{1}$ and $\gamma_{2}$ to denote these two terms from here forward.  They have the following three properties that will come in handy below.  Firstly, their values are strictly positive due to the square and non-degeneracy assumption ($\text{trace}\{\tilde{\mathbb{I}}_{l}\} \ne 0$ \ $\forall \ l$).  We emphasize this by giving them the superscript $> 0$.  Secondly, we know their derivative w.r.t.~$\tau_{l}$ is zero.  This is true for $\gamma_{1}$ because it depends on $\tau_{l}$ only through $\tilde{\mathbb{I}}$ and therefore only through the sign function.  For $\gamma_{2}$, it does not depend on $\tau_{l}$; we have factored it out already.  Thirdly, $\gamma_{1}$ and $\gamma_{2}$ are bounded w.r.t.~$\tau_{l}$.  We know this in the former case because, again, $\tau_{l}$ controls only the `gates' in $\gamma_{1}$ and $\mPhi$ is bounded.  The latter case is trivial due to $\gamma_{2}$ not being a function of $\tau_{l}$.  Now substituting back into Equation \ref{expKL_1}, we have \begin{equation}\begin{split}
    \mathbb{E}_{\rmW_{l} | \tau_{l}}[(\rf^{+}_{out})^{2}]  - 2 \mathbb{E}_{\rmW_{l} | \tau_{l}}[\rf^{+}_{out}] & f_{out}^{0} +  (f_{out}^{0})^{2} \\ &= \gamma_{1}^{\scriptscriptstyle{> 0}} + \tau_{l} \gamma_{2}^{\scriptscriptstyle{> 0}}  - 2f_{out}^{0} f_{out}^{0} + (f_{out}^{0})^{2} \\ &= \gamma_{1}^{\scriptscriptstyle{> 0}} + \tau_{l} \gamma_{2}^{\scriptscriptstyle{> 0}}  - (f_{out}^{0})^{2}.
\end{split}\end{equation}  

All that is left to do is to check for injectivity and surjectivity.  For the former, we need to verify the derivative has a constant sign, and we find that: \begin{equation*}
  \frac{\partial}{\partial \tau_{l}} \left( \mathbb{E}_{\rmW_{l} | \tau_{l}}[(\rf^{+}_{out})^{2}]  - 2 \mathbb{E}_{\rmW_{l} | \tau_{l}}[\rf^{+}_{out}] f_{out}^{0} +  (f_{out}^{0})^{2} \right) = \gamma_{2}^{\scriptscriptstyle{> 0}}.
\end{equation*} Clearly, the derivative is always positive, meaning the function is strictly increasing.  Lastly, for surjectivity, it is sufficient to check that \begin{equation*}\begin{split}
    \gamma_{1}^{\scriptscriptstyle{> 0}} + \tau_{l} \gamma_{2}^{\scriptscriptstyle{> 0}}  - (f_{out}^{0})^{2} \rightarrow 0^{+} \ \ \ &\text{ as } \ \ \ \tau_{l} \rightarrow 0^{+} \\
    \gamma_{1}^{\scriptscriptstyle{> 0}} + \tau_{l} \gamma_{2}^{\scriptscriptstyle{> 0}}  -  (f_{out}^{0})^{2} \rightarrow \infty \ \ \ &\text{ as } \ \ \ \tau_{l} \rightarrow \infty.
\end{split}
\end{equation*}  As $\tau \rightarrow 0$, we have that $\gamma_{1} \rightarrow (f^{0}_{out})^{2}$, therefore causing the first and last terms to cancel.  The remaining $\tau_{l} \gamma_{2}^{\scriptscriptstyle{> 0}}$ term is then forced to zero simply by $\tau$ going to zero.  As $\tau \rightarrow \infty$, the second term is sufficient for the full quantity to go to infinity since all other terms are bounded as a function of $\tau$. 

\subsection{Categorical Observation Model}
In the previous subsection, we showed that bijectivity is preserved at least up through the neural network's linear output $\rmF_{out}^{+}$.  We now consider the case of a categorical observation model.  This case is harder to show explicitly because the expectation cannot be pushed through to individual terms as we did above.  Rather, we base the argument on the fact that compositions of bijections form a bijection.  In particular, given that $f_{out}^{+}$ is bijective, we need the three following operations to preserve that bijectivity: $\mathbb{E}_{\rmXi} \circ  \text{KLD} \circ \text{\texttt{softmax}} \circ{f_{out}^{+}}(\tau_{l})$.  

First considering the softmax, the softmax function is defined as: 
\begin{equation*}
\pi_{j} = \frac{\exp\{\tau x_{j}\}}{\sum_{d=1}^{D} \exp\{\tau x_{d}\}}
\end{equation*} where $x_{j} \in \mathbb{R}$, $\tau \in \mathbb{R}^{+}$, $0<\pi_{j}<1$, and $\sum_{d=1}^{D} \pi_{d} = 1$.  We want to prove that the softmax is invertible w.r.t.~the scale $\tau$ for fixed $\vx = \{x_{1},\ldots,x_{D}\}$.  Since $\tau$ is a scalar, it suffices that only one $\pi_{j}$ need be invertible.  Below we show bijectity by showing injectivity and then surjectivity.  We can show injectivity by examining the softmax's derivative w.r.t.~$\tau$ and showing it has a constant sign: \begin{equation*}
    \pi'_{j}(\tau) = \underbrace{\pi_{j}}_{(+)} \underbrace{\left(x_{j} - \sum_{d=1}^{D} \pi_{d} x_{d}\right)}_{(+) \ \text{ or } \ (-)}
\end{equation*} where the first term $\pi_{j}$ is positive and the second could be positive or negative, depending on the $\vx$ values.  Denoting the maximum element of $\vx$ as $x_{\text{max}} = \max_{d} x_{d}$, the derivative in this dimension is: \begin{equation*}
     \pi'_{\text{max}}(\tau) = \underbrace{\pi_{\text{max}}}_{(+)} \underbrace{\left(x_{\text{max}} - \sum_{d=1}^{D} \pi_{d} x_{d}\right)}_{(+)} > 0,
\end{equation*} and now we know the second term is strictly positive since $\sum_{d=1}^{D} \pi_{d} x_{d}$ is a convex combination of $\vx$.  That is, assuming that there's at least one $x_{d} < x_{\text{max}}$, then $\sum_{d=1}^{D} \pi_{d} x_{d} < x_{\text{max}}$.  Conversely, for $x_{\text{min}} = \min_{d} x_{d}$, we have: 
\begin{equation*}
     \pi'_{\text{min}}(\tau) = \underbrace{\pi_{\text{min}}}_{(+)} \underbrace{\left(x_{\text{min}} - \sum_{d=1}^{D} \pi_{d} x_{d}\right)}_{(-)} < 0,
\end{equation*} \begin{wrapfigure}{r}{0.2\columnwidth}
    \includegraphics[width=0.2\columnwidth]{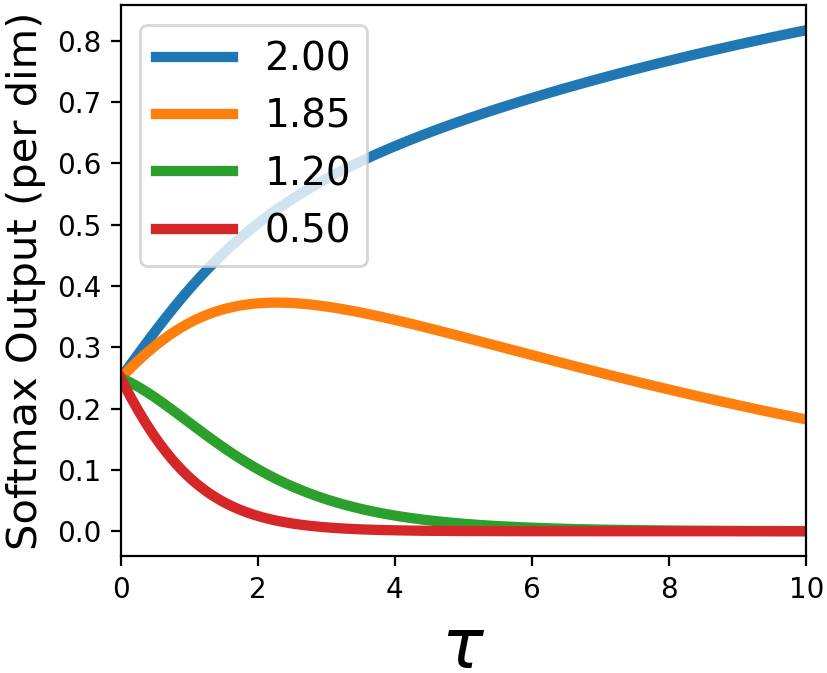}
\end{wrapfigure} since $\sum_{d=1}^{D} \pi_{d} x_{d} > x_{\text{min}}$.  Thus, we have that at least two dimensions---those corresponding to $x_{\text{min}}$ and $x_{\text{max}}$---of the softmax are injective w.r.t.~$\tau$.  
For surjectivity, it is sufficient to check the limits.  For $\tau \rightarrow 0^{+}$, $\pi_{j} \longrightarrow 1/D$ ($\forall j$), and for $\tau \rightarrow \infty$, we have $\pi_{\text{min}} \longrightarrow 0$ and $\pi_{\text{max}} \longrightarrow 1$.  Thus, $\pi_{\text{min}}$ is strictly \emph{decreasing} on $(1/D, 0)$ and that $\pi_{\text{max}}$ is strictly \emph{increasing} on $(1/D, 1)$.  In turn, \texttt{Softmax}:$\tau \mapsto \pi_{\text{min} / \text{max}}$ is a bijection.  The figure to the right shows the softmax outputs for one particular setting of $\vx$; each line represents an output $\pi_{\text{j}}$.  We see that the blue ($x=2.00)$ and red ($x=0.50$) lines, the max and min dimensions respectively, are strictly monotonic.  On the other hand, the orange line ($x=1.85$) is clearly not bijective since it doesn't pass the horizontal line test. 

On to the KLD, $\text{KL}[p^{+} || p^{0}]$ is strictly convex w.r.t.~$p^{+}$ for fixed $p^{0}$.  This is indeed our setting since only $p^{+}$ is a function of $\tau_{l}$.  This strict convexity implies that $\text{KL}[p^{+} || p^{0}]$ has a unique, global minimum at $\tau = 0$ and is strictly increasing as $\tau \rightarrow \infty$---and hence, is bijective.  Lastly, the expectation does not have an analytical solution, and thus here (as well as in practice), we consider the Monte Carlo approximation: \begin{equation*}
    \mathbb{E}_{\rmXi_{l}}\text{KL}\left[ p(\rvy | \rmX, \rmXi, \tau_{l}) || p_{0}(\rvy | \rmX) \right] \approx \frac{1}{S} \sum_{s=1}^{S} \text{KL}\left[ p(\rvy | \rmX, \hat{\mXi}_{s}, \tau_{l}) || p_{0}(\rvy | \rmX) \right]
\end{equation*} where the approximation becomes exact as $S \rightarrow \infty$.  The Monte Carlo approximation is a weighted sum of bijective, strictly increasing functions and therefore is also strictly increasing and bijective.   


\subsection{Residual Networks}
The preceding two sections hold as well for residual networks.  In fact, residual networks allow us to relax the non-degeneracy assumption.  That is, for invertibility w.r.t.~$\tau_{l}$, we can have $\vh_{j}=\mathbf{0}$ for $j > l$ just so long as there is a path of skip (identity) connections from layer $l$ to layer $m > j$.  This identity path preserves the information about $\tau_{l}$ that would be lost due to $\vh_{j}=\mathbf{0}$.

\section{DRAWING SAMPLES FROM THE PREDCP}
The user may want to draw samples from the PredCP---for instance, to perform a prior predictive check \citep{box1980sampling}.  In general, sampling from a reparameterized model can be done via: $\hat{x} \sim p(\rx)$, $\hat{\theta} = T(\hat{x})$ where $p(\rx)$ is the base distribution and $T$ is the transformation \citep{papamakarios2019normalizing}.  Applying this formula to the PredCP, in theory, we could draw samples via: \begin{equation}
    \hat{\kappa} \sim \pi(\rkappa), \ \ \ \hat{\tau} = T^{-1}(\hat{\kappa})
\end{equation} where $\hat{\kappa}$ represents a sampled value of $\mathbb{E}_{\rvtheta | \rtau}\text{KL}\left[ p_{+} || p_{0} \right]$ and $T^{-1}$ represents an inversion of the expected-KLD, yielding $\rtau$ as a function of $\rkappa$.  Unfortunately, the analytical inverse of $\mathbb{E}_{\rvtheta | \rtau}\text{KL}\left[ p_{+} || p_{0} \right]$ is not available in general.  Instead, we use the numerical inversion technique proposed by \citet{song2019mintnet}.  See Algorithm \ref{sampling} below.
\begin{algorithm}
\begin{algorithmic}
\STATE \textbf{Input}: Prior $\pi(\rkappa)$, \ number of iterations $T$, \ step size $\alpha$\\
\textbf{Sample} \ \  $\hat{\kappa} \sim \pi(\rkappa)$ \\
\textbf{Initialize} \ \  $\tau_{0} \leftarrow \hat{\kappa}$ \\
\textbf{For} \ $t \ = \ [1, T]$: \\
\hspace{1.5em} \textbf{Compute} \ \ $\mathbb{D}(\tau_{t-1}) = \mathbb{E}_{\rvtheta | \tau_{t-1}}\text{KL}\left[ p_{+} || p_{0} \right]$\\
\hspace{1.5em} \textbf{Compute} \ \ $\partial \mathbb{D}(\tau_{t-1}) / \partial \tau_{t-1} $\\
\hspace{1.5em} \textbf{Update} \ \ $\tau_{t} \leftarrow \tau_{t-1} - \alpha \left(  \partial \mathbb{D}(\tau_{t-1}) / \partial \tau_{t-1} \right)^{-1} \left( \mathbb{D}(\tau_{t-1}) - \hat{\kappa} \right) $\\
\textbf{Return} \ \ $\tau_{T}$
\end{algorithmic}
\caption{Sampling from the PredCP}
\label{sampling}
\end{algorithm}

\begin{figure*}
\centering
\includegraphics[width=\linewidth]{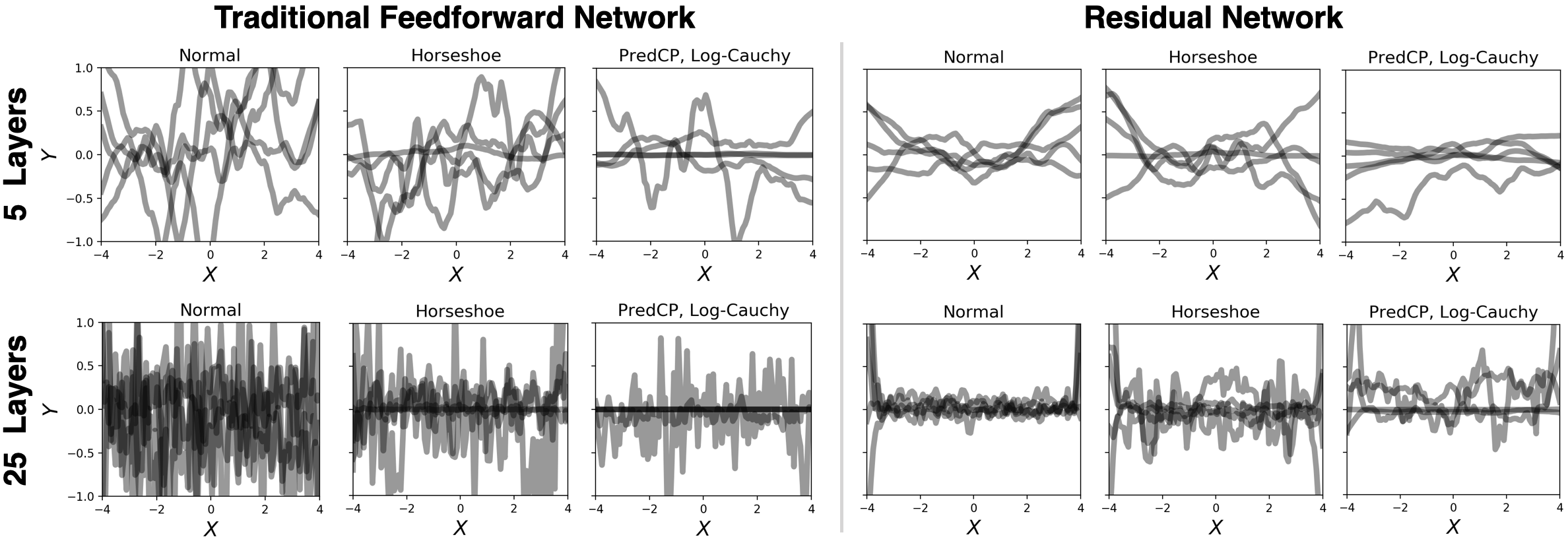}
\caption{\textit{Induced Prior Over Functions}.  We compare a standard normal prior, the horseshoe ($\rtau \sim C^{+}(0,1)$), and the depth-wise PredCP ($\pi(\rkappa) = \text{Log-Cauchy}(0,1)$) for $5$ and $25$ layer NNs.
}
\label{fig:func_priors}
\end{figure*}

We used Algorithm \ref{sampling} to sample from the depth-wise PredCP and ancestral sampling to draw functions from the NN.  We considered both residual and non-residual architectures, both having batch normalization applied at each hidden layer.  The observation model is Gaussian with $\sigma_{y}=1$.  We used a log-Cauchy$(0,1)$ KLD prior, $\alpha=5 \times 10^{-5}$, $T=20$, and $5$ Monte Carlo samples for the $\rvtheta | \rtau$ expectation.  In Figure \ref{fig:func_priors}, we show $5$ function samples for the standard Normal, horseshoe ($\rtau \sim C^{+}(0,1)$), and PredCP for $5$ and $25$ hidden layers.  The PredCP's samples are notably simpler.    

\section{MODULAR PRIOR SPECIFICATION FOR META-LEARNING}
\label{app:prior_terms}
We provide details for specifying and evaluating \textit{modular} priors for meta-learning.
Following \citet{chen2019modular}, we split $\rvtheta$ into $M$ distinct modules (e.g., layers), such that $\rvtheta = [\rvtheta_1^T, \hdots, \rvtheta_M^T]^T$.
Our goal is then to place a separate prior on each module, and allow the modules to adapt differently to new tasks.
Denoting $\vphi_m$ and $\rtau_m$ as the global parameters and shrinkage parameter of module $m$, respectively, we have that $\rvtheta_{t,m} \sim \text{N} (\vphi_m, \rtau_m \mathbb{I} ) $ for the local parameters of task $t$ at module $m$.
By specifying $p_m(\rtau_m)$, we can control how much each module is allowed to deviate from the global model.
For example, \citet{chen2019modular} place an improper flat prior on $\rtau_m$ and perform MAP estimation. 
Importantly, to perform MAP estimation for $\{\rtau_m\}_{m=1}^M$, we need only evaluate the log density of the priors $p_m(\rtau_m)$, and add these terms to the outer-loop optimization objective \citep{chen2019modular}.

\paragraph{Defining a PredCP Prior for $\rtau$}
In this setting, It is natural to consider the global parameters as defining the base model for the PredCP prior.
We can achieve this by specifying the prior to be $p_{0, m}(\rvtheta_{t, m}) = \delta(\rvtheta_{t, m} - \vphi_m)$.
Denoting $p_0(\rvtheta) = \prod_{m=1}^M p_{0, m}(\rvtheta_m)$, we then have
\[
    p_0(\rvy | \vx) = \int p(\rvy | \vx, \rvtheta) p_0(\rvtheta) \mathrm{d}\rvtheta = p_\vphi(\rvy | \vx),
\]
where we denote $p_\vphi(\rvy | \vx)$ as the predictions made by the model using only global parameters.
Next, for every module $m$, we can define the extended model 
\[
    p_m(\rvy | \vx, \rtau_m) = \int p(\rvy | \vx, \vphi, \rvtheta_m) p_m(\rvtheta_m | \rtau_m) \mathrm{d} \rvtheta_m,
\]
where we use the notation $p(\rvy | \vx, \vphi, \rvtheta_m)$ to denote the model that uses $\vphi$ for all but module $m$, which uses $\rvtheta_m$.
Using the notation from the main text, we further denote $p_0(D_t) = \prod_{\vx, \rvy \in D_t} p_0(\rvy | \vx)$ and $p_m(D_t | \rvtheta_{t, m}) = \prod_{\vx, \rvy \in D_t} p_m( \rvy | \vx, \vphi, \rvtheta_m)$.
Recall that using the KLD upper bound, the PredCP prior for this setting can be expressed as
\[
    p(\rtau_m) =  \pi\left( \frac{1}{T} \sum_{t=1}^{T} \mathbb{E}_{\rvtheta_{t, m} | \rtau_m}\text{KL}\left[  p_{m}(\mathcal{D}_{t} | \rvtheta_{t, m}) \ || \   p_{0}(\mathcal{D}_{t}) \right]  \right) \ \left| \frac{1}{T} \sum_{t} \frac{\partial  \  \mathbb{E}_{\rvtheta_{t} | \rtau}\text{KL}_{t}}{\partial \ \rtau} \right|.
\]
We can approximate the intractable term inside $\pi(\rkappa)$ without bias via Monte-Carlo sampling:
\[
    \mathbb{E}_{\rvtheta_{t, m} | \rtau_m} \text{KL}\left[  p_{m}(\mathcal{D}_{t} | \rvtheta_{t, m}) \ || \   p_{0}(\mathcal{D}_{t}) \right] 
    \approx \frac{1}{L} \sum_{l=1}^L \text{KL}\left[  p_{m}(\mathcal{D}_{t} | \rvtheta_{t, m}^{(l)}) \ || \   p_{0}(\mathcal{D}_{t}) \right],
\]
where $\rvtheta_{t, m} \sim \text{N}(\vphi_m, \rtau_m \mathbb{I})$.
Note that the KLD term itself is a sum over the KLD terms for each $(\vx, \rvy)$ in the support set.
In turn, each of these terms is a KLD between categorical likelihoods, which is easily computed.
As detailed in the main text, we divide the sum by the number of points in the set to ease the reasoning about the parameters of $\pi_{\mathrm{KL}}$. 
We can further achieve an unbiased estimator to the term inside $\pi_{\mathrm{KL}}$ using stochastic mini-batches of tasks, as is standard in the few-shot classification literature.

Finally, assuming a factorization of the $\rtau_m$'s, we can simply evaluate the log-density for each prior, and compute their sum.
In practice, we use single-sample estimators of the KL term, and compute the prior terms over batches of tasks to reduce the number of forward passes through the network required to compute the objective.
The procedure for evaluating the PredCP prior for a batch of $B$ tasks is detailed in Algorithm \ref{alg:meta-learning-evaluation}. 
Here we use the notation $\mathrm{CNN}(\vx; \cdot)$ to denote a forward pass through a convolutional neural network using a set of parameters, applied to an input.
We treat the output of such a call as the logits of a categorical distribution.

\begin{algorithm}
    \begin{algorithmic}
    \STATE \textbf{Input}: Global parameters $\{ \vphi_m \}_{m=1}^M$,  network architecture $\mathrm{CNN}$ \\
    \STATE \textbf{Input}: Prior $\pi(\rkappa)$, \ current values $\{ \rtau_m \}_{m=1}^M$\ \\
    \STATE \textbf{Input}: Inputs from all tasks in batch $\{ \{ \vx_{n, b} \}_{n=1}^N \colon b=1, \hdots, B \}$ \\
    
    \hspace{1.5em}
    
    \STATE \textbf{Compute} $p^{n,b}_0 \leftarrow \mathrm{CNN}(\vx_{n,b}; \vphi)$ \\
    \STATE \textbf{For} \ $m \ = \ [1, M]$: \\
    \STATE \hspace{1.5em} \textbf{Sample} \ \ $\rvtheta_m \sim \text{N}\left(\vphi_m, \rtau_m \mathbb{I} \right)$\\
    \STATE \hspace{1.5em} \textbf{Compute} $p^{n,b}_m \leftarrow \mathrm{CNN}(\vx_{n,b}; \vphi, \rvtheta_m)$ \\
    \STATE \hspace{1.5em} \textbf{Compute} $\mathrm{KL}_m \leftarrow \frac{1}{NB} \sum_{n,b} \mathrm{KLD}(p_m^{n,b}, p_0^{n,b})$ \\
    \STATE \hspace{1.5em} \textbf{Compute} $\log \pi_m \leftarrow \log \pi_{\mathrm{KL}} ( \mathrm{KL}_m) + \log \left| \frac{\partial \mathrm{KL}_m}{\partial \rtau_m}  \right|$ \\
    \textbf{Return} \ \ $\sum_{m=1}^M \log \pi_m $
    \end{algorithmic}
\caption{Single sample evaluation of $\pi_\rtau$ for modular meta-learning}
\label{alg:meta-learning-evaluation}
\end{algorithm}

\section{LOGISTIC REGRESSION EXPERIMENTAL DETAILS}\label{app:exp1}  For the logistic regression experiment in Section 6 (Table 1), we used the probabilistic programming language \texttt{Stan} \citep{carpenter2017stan} for the implementation of both Markov chain Monte Carlo (MCMC) and variational inference (VI) \citep{kucukelbir2017automatic}.  In both cases, we performed inference for the full posterior $p(\rvbeta, \rvlambda, \rtau | \mX, \vy)$.  Following \citet{piironen2017hyperprior}'s implementation, we used a non-centered parameterization: $\rbeta_{d} = \lambda_{d} \cdot \tau \cdot \xi_{d} $, $\xi_{d} \sim \text{N}(0,1)$.  For MCMC, we left \texttt{Stan} at its default settings.  We evaluated the predictive log-likelihood using $4000$ posterior samples (the default output).  For VI, \texttt{Stan} uses a mean-field Normal posterior, appropriately transforming all variables so that their support is $\mathbb{R}$.  Again we left all hyper-parameters at their defaults.  The predictive log-likelihood was evaluated with $1000$ samples drawn from the approximate posterior.  For the PredCP, we used $10$ samples to evaluate the Monte Carlo expectation over $\rvtheta | \rtau$ for \texttt{colon} and \texttt{breast}.  For \texttt{allaml}, we used only one Monte Carlo sample due to the data set being larger and requiring more time to run the MCMC.  The \texttt{allaml} and \texttt{colon} data sets were downloaded from \url{http://featureselection.asu.edu/datasets.php}.  \texttt{breast} was downloaded from \url{https://archive.ics.uci.edu/ml/datasets/Breast+Cancer+Coimbra}.  We standardized the features using the z-transform $(x - \hat{\mu}) / \hat{\sigma}$, which we found to improve the speed at which the MCMC converged.  We made $20$ $80$\% - $20$\% train-test splits for all three data sets.  We left out $10$\% of the training set for each as a validation set that we ultimately did not use.    

\section{RESNET EXPERIMENTAL DETAILS}\label{app:exp2}
For the resnet regression experiment in Section 6 (Table 2), we followed the experimental framework of \citet{nalisnick19dropout}\footnote{\url{https://github.com/enalisnick/dropout_icml2019}} (which followed \citet{gal2016dropout} and \citet{hernandez2015probabilistic}).  All networks had two hidden layers, ReLU activations, and no batch or layer normalization.  The posterior was approximated as: \begin{equation}
    p(\{ \rmW_{l}, \rvlambda_{l}, \rtau_{l} \}_{l=1}^{3} | \vy, \mX) \approx \prod_{l=1}^{3} q(\rmW_{l}) \ q(\rvlambda_{l}) \ q(\rtau_{l}) = \prod_{l=1}^{3} \text{N}(\rmW_{l}; \vmu_{l}, \text{diag}\{ \mSigma_{l} \}) \ \delta[\rvlambda_{l}] \ \delta[\rtau_{l}].
\end{equation}  The weight approximation (Bayes-by-backprop \citep{blundell2015weight}, fully factorized Gaussian) was optimized using Adam \citep{kingma2014adam} with a learning rate of $1\times 10^{-3}$ (other parameters left at Tensorflow defaults), using mini-batches of size $32$, and run for $4500$ epochs.  The Monte Carlo expectations in the ELBO and PredCP both used $10$ samples and \textit{flipout} \citep{wen2018flipout} for decorrelation.  The ARD scales $\rvlambda$ could be updated in closed-form for all models.  The PredCP does not allow for a closed-form $\rtau$ update (for ADD) and so we used an iterative maximization step.  We used only one step per update so that the PredCP's training time was comparable to the other models'.  The UCI data sets were standardized and divided into 20 $90$\% - $10$\% train-test splits, following \citet{hernandez2015probabilistic}.  The test set RMSE was calculated using $500$ samples from the $\text{N}(\rmW_{l})$ posterior.  For the fixed scale model, we selected the better performing of $\tau_{0} = \{.1, 1\}$.  For the PredCP, we used the log-Cauchy$(0,1)$ KLD prior, finding it worked well in the logistic regression experiment and generated sensible sample functions.  In Algorithm \ref{depthPredCP}, we provide pseudo-code for evaluating the depth-wise (log) PredCP.
\begin{algorithm}
\begin{algorithmic}
\STATE \textbf{Input}: Scales $\rtau_{1},\ldots,\rtau_{L}$, prior $\pi(\rkappa)$, \ number of Monte Carlo samples $S$, \ feature matrix $\mX$\\
\textbf{Initialize} \ \  $\pi \leftarrow 0$ \\
\textbf{Initialize} \ \  $p_{0} \leftarrow p(\rvy | \mX, \rmW_{\text{in}}, \rmW_{\text{out}})$ \\
\textbf{For} \ $l \ = \ [1, L]$: \\
\hspace{1.5em} \textbf{Sample} \ \ $\hat{\mW}_{l,1}, \ldots, \hat{\mW}_{l,S} \sim p(\rmW_{l} | \rtau_{l})$\\
\hspace{1.5em} \textbf{For} \ $s \ = \ [1, S]$: \\
\hspace{3em} \textbf{Compute} \ \ $p_{+,s} = p(\rvy | \mX, \rmW_{\text{in}}, \{\hat{\mW}_{j, s} \}_{j=1}^{l}, \rmW_{\text{out}})$\\
\hspace{1.5em} \textbf{Compute} \ \ $\hat{\kappa}_{l} = \frac{1}{S} \sum_{s=1}^{S} \text{KL}[p_{+, s} || p_{0, s}]$\\
\hspace{1.5em} \textbf{Update} \ \ $\pi \leftarrow \pi + \log \pi(\hat{\kappa}_{l}) + \log \left| \partial \hat{\kappa}_{l} / \partial \tau_{l} \right|$\\
\hspace{1.5em} \textbf{Update} \ \ $\{p_{0,1}, \ldots, p_{0,S} \}  \leftarrow  \{p_{+,1}, \ldots, p_{+,S} \} $\\
\textbf{Return} \ \ $\pi$  \ \ (\textit{log-PredCP})
\end{algorithmic}
\caption{Evaluating the Depth-Wise Log-PredCP}
\label{depthPredCP}
\end{algorithm}

\section{FEW-SHOT CLASSIFICATION: EXPERIMENTAL DETAILS AND ADDITIONAL RESULTS}\label{app:exp3}
\label{app:few_shot_additional_results}
We provide experimental details and results for our few-shot classification experiments.

\subsection{Data Details}
We use two standard few-shot classification benchmarks for our experiments: mini-ImageNet \citep{vinyals2016matching} and few-shot CIFAR100 (FC; \citep{oreshkin2018task}).
For mini-ImageNet, images are first down-sampled to 84x84, and then normalized.
For FC, the images are classified in their 32x32 format after normalization.
We use the standard split as suggested by \citet{vinyals2016matching} for mini-ImageNet, containing 64 training classes, 16 validation classes, and 20 test classes.
For FC, we follow the protocol proposed by \citet{oreshkin2018task}, using the CIFAR100 super-classes to split the data. See \citet{oreshkin2018task} supplementary for full details.
An $N-way$, $K-shot$ task is randomly sampled according to the following procedure:
\begin{itemize}
    \item Sample $N$ classes from the appropriate set uniformly at random.
    \item For each class, sample $K$ examples uniformly at random for the context / support set.
    \item For each class, sample $15$ examples uniformly at random for the target / query set.
\end{itemize}
Evaluation is conducted by randomly sampling 600 tasks from the test set.
Average accuracy and standard errors are reported.

\subsection{Network Architectures}
For all tasks, we use the standard convolutional architecture proposed by \citet{finn2017model}.
Each network is comprised of four convolutional blocks, followed by a linear classier.
For mini-ImageNet, we use a standard 3x3 convolution with 32 channels, followed by a max-pool (stride 2), a ReLU non-linearity, and a batch normalization layer.
We flatten the output of the final layer, leading to an 800d representation, which is then passed through the linear classifier.

For FC, we employ the same architecture, but with 64 channels, and no max-pooling.
A global-average pooling is applied to the output of the final convolutional layer, resulting in a 64d representation, which is then passed through the linear classifier.

We used the standard hyper-parameters proposed by \citet{finn2017model}, without any tuning.
In particular, we use 5 gradient steps for the inner loop during training, and 10 at test time.
The inner learning rate is fixed to 0.01, and the meta-learning rate is 1e-3. 
We use a meta-batch size of 4, and train all models for 60,000 iterations.

\subsection{Hyper-Priors and Additional Results}
\label{app:hyper-priors}
\begin{table}[htb]
\scriptsize
\caption{Complete results for few-shot classification with few-shot CIFAR100. Considering four priors for $\sigma$MAML and four base priors for PredCP.}
\label{tab:fs_cifar100}
\begin{tabular}{llclclclc}
\toprule
Beta & \multicolumn{2}{c}{Half-Cauchy}               & \multicolumn{2}{c}{Log-Cauchy}                    & \multicolumn{2}{c}{Gem}          & \multicolumn{2}{c}{Exponential}                                                             \\ 
        & 1-shot       & 5-shot      & 1-shot            & 5-shot     & 1-shot     & 5-shot     & 1-shot      & 5-shot  \\ 
\midrule
\textsc{PredCP} & & & & & & & & \\
1e-0 & 33.2 $\pm$ 1.8 & 51.7 $\pm$ 0.9          & 37.7 $\pm$ 1.7          & 51.8 $\pm$ 0.9          & 37.9 $\pm$ 1.9 & 52.0 $\pm$ 0.97 & 38.9 $\pm$ 1.8          & 51.4 $\pm$ 1.0 \\
1e-1 & 39.6 $\pm$ 1.6 & 51.4 $\pm$ 0.9          & 37.9 $\pm$ 1.9          & 50.6 $\pm$ 0.7          & 40.9 $\pm$ 1.8 & 51.7 $\pm$ 0.9  & 39.5 $\pm$ 1.9          & 51.1 $\pm$ 0.8 \\
1e-2 & 38.2 $\pm$ 1.8 & 50.9 $\pm$ 0.8          & 40.1 $\pm$ 1.9          & 52.5 $\pm$ 0.9          & 37.8 $\pm$ 1.8 & 51.6 $\pm$ 0.8  & 40.3 $\pm$ 1.9          & 49.1 $\pm$ 1.1 \\
1e-3 & 39.7 $\pm$ 1.9 & \textbf{52.9 $\pm$ 0.9} & 40.2 $\pm$ 1.8          & 50.6 $\pm$ 1.0          & 38.8 $\pm$ 1.8 & 51.3 $\pm$ 0.9  & 38.5 $\pm$ 1.8          & 50.8 $\pm$ 0.9 \\
1e-4 & 37.4 $\pm$ 1.7 & 52.7 $\pm$ 0.9          & 37.7 $\pm$ 1.8          & 50.5 $\pm$ 0.9          & 35.8 $\pm$ 1.7 & 51.5 $\pm$ 0.9  & \textbf{41.2 $\pm$ 1.8} & 50.9 $\pm$ 0.9 \\ 
\midrule
\textsc{Shrinkage Priors} & & & & & & & & \\
1e-0 & 36.9 $\pm$ 1.8 & 51.4 $\pm$ 0.9          & 39.5 $\pm$ 1.9          & 52.2 $\pm$ 1.1          & 39.8 $\pm$ 1.8 & 52.4 $\pm$ 0.9  & 39.4 $\pm$ 1.8          & 52.6 $\pm$ 0.7 \\
1e-1 & 39.8 $\pm$ 1.9 & \textbf{52.7 $\pm$ 0.8} & \textbf{40.9 $\pm$ 1.9} & \textbf{52.7 $\pm$ 0.9} & 36.6 $\pm$ 1.8 & 52.5 $\pm$ 0.8  & 38.7 $\pm$ 1.9          & 51.6 $\pm$ 1.1 \\
1e-2 & 38.2 $\pm$ 1.8 & 52.0 $\pm$ 0.9          & 38.8 $\pm$ 1.7          & 51.2 $\pm$ 0.8          & 36.4 $\pm$ 1.8 & 51.4 $\pm$ 0.9  & 39.1 $\pm$ 1.9          & 52.2 $\pm$ 0.9 \\
1e-3 & 39.3 $\pm$ 1.7 & 51.3 $\pm$ 0.7          & 35.9 $\pm$ 1.8          & 50.8 $\pm$ 0.9          & 40.8 $\pm$ 1.9 & 51.9 $\pm$ 1.0  & 37.5 $\pm$ 1.9          & 52.2 $\pm$ 0.8 \\
1e-4 & 38.6 $\pm$ 1.7 & 51.5 $\pm$ 0.9          & 40.9 $\pm$ 1.8          & 50.1 $\pm$ 0.9          & 38.8 $\pm$ 1.8 & 51.8 $\pm$ 0.9  & 40.1 $\pm$ 1.8          & 51.4 $\pm$ 0.9 \\ 
\midrule 
MAML      (flat $\theta_t$ prior)  & 35.6 $\pm$ 1.8 & 50.3 $\pm$ 0.9  & & & & & & \\ 
$\sigma$MAML (flat $\tau$ prior) & 39.3 $\pm$ 1.8 & 51.0 $\pm$ 1.0  & & & & & & \\ 
\bottomrule
\end{tabular}
\end{table}
\begin{table}[htb]
\scriptsize
\caption{Complete results for few-shot classification with mini-ImageNet. Considering four priors for $\sigma$MAML and four base priors for PredCP.}
\label{tab:fs_mini_imagenet}
\begin{tabular}{lcccccccc}
\toprule
\multirow{2}{*}{Beta} & \multicolumn{2}{c}{Half-Cauchy} & \multicolumn{2}{c}{log-Cauchy} & \multicolumn{2}{c}{GEM} & \multicolumn{2}{c}{Exponential} \\
                      & 1-shot       & 5-shot      & 1-shot            & 5-shot     & 1-shot     & 5-shot     & 1-shot      & 5-shot  \\ 
\midrule
\textsc{PredCP} & & & & & & & & \\
1e-0                  & 33.4 $\pm$ 1.7  & 60.7 $\pm$ 0.8  & 28.6 $\pm$ 1.6          & 59.8 $\pm$ 0.8  & 28.7 $\pm$ 1.6  & 59.1 $\pm$ 0.8  & 30.2 $\pm$ 1.7  & 59.1 $\pm$ 0.8 \\
1e-1                  & 47.9 $\pm$ 1.7  & 60.4 $\pm$ 0.9  & \textbf{49.3} $\pm$ 1.7 & 60.4 $\pm$ 0.7  & 47.5 $\pm$ 1.9  & 61.4 $\pm$ 0.8  & 47.7 $\pm$ 1.7  & 60.6 $\pm$ 0.9 \\
1e-2                  & 47.0 $\pm$ 1.8  & 60.3 $\pm$ 0.7  & 45.9 $\pm$ 1.6          & 60.7 $\pm$ 0.8  & 46.7 $\pm$ 1.8  & 61.7 $\pm$ 0.7  & 47.2 $\pm$ 1.9  & \textbf{61.9} $\pm$ 0.9 \\
1e-3                  & 46.4 $\pm$ 1.7  & 61.3 $\pm$ 0.8  & 48.1 $\pm$ 1.8          & 61.2 $\pm$ 0.9  & 47.9 $\pm$ 1.7  & 60.5 $\pm$ 0.8  & 46.9 $\pm$ 1.7  & 60.5 $\pm$ 0.8  \\
1e-4                  & 47.7 $\pm$ 1.8  & 60.4 $\pm$ 0.9  & 47.7 $\pm$ 1.8          & 60.3 $\pm$ 0.9  & 48.1 $\pm$ 1.9  & 60.1 $\pm$ 0.9  & 48.3 $\pm$ 1.8  & 60.4 $\pm$ 0.9  \\ 
\midrule
\textsc{Shrinkage Priors} & & & & & & & & \\
1e-0                  & 42.9 $\pm$ 1.8  & 57.3 $\pm$ 0.9 & \textbf{48.5} $\pm$ 1.7 & \textbf{60.9} $\pm$ 0.9 & 24.7 $\pm$ 1.5 & 29.0 $\pm$ 0.7 & 44.5 $\pm$ 1.8 & 57.8 $\pm$ 0.9 \\
1e-1                  & 46.5 $\pm$ 1.8  & 59.2 $\pm$ 0.9 & 47.2 $\pm$ 1.8          & 59.7 $\pm$ 0.9          & 45.8 $\pm$ 1.8 & 58.7 $\pm$ 0.8 & 47.0 $\pm$ 1.8 & 59.6 $\pm$ 0.9 \\
1e-2                  & 46.8 $\pm$ 1.7  & 59.3 $\pm$ 0.8 & 46.3 $\pm$ 1.9          & 59.3 $\pm$ 0.8          & 46.7 $\pm$ 1.7 & 60.1 $\pm$ 0.7 & 46.8 $\pm$ 1.8 & 59.1 $\pm$ 0.9 \\
1e-3                  & 47.3 $\pm$ 1.9  & 59.2 $\pm$ 0.9 & 47.7 $\pm$ 1.7          & 59.3 $\pm$ 0.9          & 47.2 $\pm$ 1.9 & 60.1 $\pm$ 0.9 & 47.5 $\pm$ 1.9 & 59.5 $\pm$ 1.0 \\
1e-4                  & 46.7 $\pm$ 1.8  & 59.2 $\pm$ 0.9 & 47.4 $\pm$ 1.7          & 60.2 $\pm$ 0.7          & 48.0 $\pm$ 1.5 & 59.5 $\pm$ 0.9 & 47.8 $\pm$ 1.7 & 58.6 $\pm$ 0.9 \\ 
\midrule
MAML (flat $\theta_{t}$ prior)       & 45.6 $\pm$ 1.8 & 58.4 $\pm$ 0.9 & & & & & & \\
$\sigma$MAML (flat $\tau$ prior) & 47.4 $\pm$ 1.8 & 60.1 $\pm$ 0.9 & & & & & & \\
\bottomrule
\end{tabular}
\end{table}
For the modular shrinkage model, we experiment with four hyper-priors: Half-Cauchy, log-Cauchy, a mixture of Gamma and Exponential (GEM), and standard Exponential.
Each of these is experimented with as a standard shrinkage prior, as well as the base prior for the PredCP.
The parameters of the hyper-priors are fixed throughout experimentation, and are given as follows:
\begin{itemize}
    \item Half-Cauchy: $p(\sigma), \pi_{\mathrm{KL}} = \text{half-Cauchy}(1.0)$
    \item Log-Cauchy: $p(\sigma), \pi_{\mathrm{KL}} = \text{log-Cauchy}(2.0)$
    \item GEM: $p(\sigma), \pi_{\mathrm{KL}} = 0.5 \left(\Gamma(0.2, 2.0) + \text{ Exp}(0.5) \right)$
    \item Exponetial: $p(\sigma), \pi_{\mathrm{KL}} = \text{Exp}(0.5)$
\end{itemize}
Additionally, we experimented with a vague (flat) prior, which recovers the model proposed by \citet{chen2019modular}.
In addition, we also add a constant weight $\beta$, which multiplies the prior on $\sigma$ in the outer-loop objective for both the regular shrinkage and PredCP.
For each prior, we experiment with $\beta \in \{1e-0, 1e-1, 1e-2, 1e-3, 1e-4\}$.

Tables \ref{tab:fs_cifar100} and \ref{tab:fs_mini_imagenet} provide our complete results for these experiments.
For both datasets, and for both 1- and 5- shot, we observe that a PredCP prior with appropriate weighting on the prior term provides the best performance in terms of accuracy on the test set.


\end{document}